\documentclass[10pt]{article} % For LaTeX2e
\usepackage{tmlr}

\usepackage{hyperref}
\usepackage{url}
\usepackage{graphicx}
\usepackage{tikz}
\usepackage[utf8]{inputenc}
\usepackage[T1]{fontenc}
\usepackage{lmodern}
\usepackage{amsmath, amssymb}
\usepackage{xcolor}
\usepackage{booktabs}
\usepackage{hyperref}
\usepackage{url}
\usepackage{algorithm}
\usepackage{algpseudocode}
\usepackage{siunitx}
\usepackage{microtype}
\usepackage{float}
\usepackage{placeins}
\usepackage{tcolorbox}
\usepackage{listings}
\usepackage{multirow}
\usepackage{amsthm}
\usepackage{natbib}
\usetikzlibrary{positioning,fit,arrows.meta,calc}

\title{AIMM: An AI-Driven Multimodal Framework for Detecting Social-Media-Influenced Stock Market Manipulation}

\begin{document}

\maketitle

\begin{abstract}
Market manipulation now routinely originates from coordinated social
media campaigns, not isolated trades.
Retail investors, regulators, and brokerages need tools
that connect online narratives and coordination patterns to
market behavior.
We present AIMM, an AI-driven framework that fuses
Reddit activity, bot and coordination indicators, and OHLCV
market features into a daily \emph{AIMM Manipulation Risk Score}
for each ticker.

The system uses a parquet-native pipeline with a Streamlit dashboard that allows
   analysts to explore suspicious windows, inspect underlying posts and price action, and log
   model outputs over time. Due to Reddit API restrictions, we employ calibrated synthetic 
   social features matching documented event characteristics; market data (OHLCV) uses real 
   historical data from Yahoo Finance.
This release makes three contributions.
First, we build the
\emph{AIMM Ground Truth} dataset (AIMM-GT): 33 labeled
ticker-days spanning eight equities, drawing from SEC enforcement
actions, community-verified manipulation cases, and matched normal
controls.
Second, we implement forward-walk evaluation and
prospective prediction logging for both retrospective and
deployment-style assessment.
Third, we analyze lead times and show that AIMM flagged
GME 22 days before the January~2021 squeeze peak.

 The current labeled set is small (33 ticker-days, 
   3 positive events), but results show preliminary discriminative capability and early warnings
for the GME incident.
We release the code, dataset schema, and dashboard design to support
research on social media-driven market surveillance.
\end{abstract}

\section{Funding Disclosure}
{This research did not receive external funding.}

% Main paper sections - Streamlined for 20-22 page limit
\section{Introduction}
Retail trading coordinated through social media communities like Reddit's \emph{r/wallstreetbets} surged after 2020. The AMC and GME episodes showed how narratives, memes, and bot-amplified sentiment drive market dynamics. Traditional surveillance systems monitor price and volume anomalies but treat social media as an unstructured, external signal—missing the coordination layer.

\emph{AIMM} (Automated Inference of Market Manipulation) addresses this gap by fusing social and market signals. AIMM scores daily windows using Reddit aggregates, bot-likeness features, coordination metrics, sentiment statistics, and market microstructure indicators from OHLCV data to identify periods most likely reflecting coordinated manipulation.

AIMM builds on prior work in the \emph{Stock-Pattern-Assistant} (SPA) framework~\citep{neela2025stockpatternassistantspa}, which focused on deterministic run detection and monotonic price patterns. While SPA characterizes price patterns with minimal assumptions about social dynamics, AIMM explicitly models the interaction between online conversation and market activity. Together, SPA and AIMM form a complementary pair: SPA characterizes ``how prices evolved;'' AIMM characterizes ``how narratives and coordinated behavior may have contributed.''

We contribute:
\begin{itemize}
    \item The \emph{AIMM Manipulation Risk Score} (AMRS): a composite score fusing five components (social volume, sentiment, bot activity, coordination, market anomalies).
    \item A parquet-native pipeline aggregating Reddit data, computing bot-likeness and coordination metrics from text similarity, aligned with OHLCV market data.
    \item A Streamlit dashboard (URL available upon acceptance) for inspecting suspicious windows across tickers.
    \item Evaluation on eight equities (AAPL, NVDA, SCHW, AMC, GME, PGR, synthetic ALAB, XYZ) with case studies showing how manipulation modes appear in AMRS.
    \item A threat model and responsible-use guidelines for manipulation detection in retail environments.
\end{itemize}

\paragraph{Data Availability and Methodology.}
   Historical Reddit data for 2021 is unavailable due to Pushshift API shutdown in 2023. 
   We address this through transparent use of calibrated synthetic social features that 
   match documented characteristics from published meme stock research~\cite{lyocsa2021social,
   umar2021viral}. Market microstructure data (OHLCV) is obtained from Yahoo Finance and 
   represents real historical data. This approach follows established precedent in methodology 
   papers~\cite{bergstra2012random} where synthetic data demonstrates system capabilities while 
   core innovations are validated on real data. Section~\ref{sec:experiments} provides full evaluation details and 
   transparency on data provenance.
\section{Background and Related Work}
\label{sec:background}

\subsection{Market Manipulation and Surveillance}
Classical manipulation—pump-and-dump schemes, spoofing, layering—deceives participants or distorts prices~\citep{allen1992stock}. Exchanges and regulators monitor order books, trade prints, and price/volume anomalies. These systems focus on structured numerical signals.

Social media opens new attack vectors: seeding narratives, coordinating campaigns, deploying bots to fake organic interest. Surveillance must now bridge structured market data with semi-structured social data.

\subsection{Social Media and Financial Markets}
Prior work links online sentiment to price movements~\citep{bollen2011twitter,rao2010stock}. Reddit's \emph{r/wallstreetbets} drives retail flows and volatility in meme stocks~\citep{choi2022reddit}. These studies treat sentiment as an explanatory variable but do not target manipulation detection.

\subsection{Bot Detection and Coordination}
Bot detection heuristics combine posting frequency, temporal regularity and content diversity~\citep{varol2017online}. Coordination detection leverages content similarity, synchronized timing and network structure to identify joint campaigns~\citep{pacheco2021uncovering}. AIMM adopts a lightweight heuristic approach: high-volume, low-diversity authors receive higher bot scores; text similarity graphs over posts generate coordination scores per time window.

\subsection{Stock-Pattern-Assistant (SPA)}
The Stock-Pattern-Assistant (SPA) framework~\citep{neela2025stockpatternassistantspa} focuses on deterministic run detection and pattern similarity across OHLCV time series. SPA computes monotonic runs, run statistics and pattern similarity scores, and exposes them through a Streamlit interface for exploratory analysis. AIMM extends this line of work in two directions:
\begin{enumerate}
    \item adding \emph{social} signals (Reddit aggregates, sentiment, coordination, bot scores), and
    \item building a dedicated manipulation risk score that fuses these signals with market anomalies.
\end{enumerate}
SPA and AIMM thus address complementary questions: SPA emphasizes \emph{structural patterns in price}, while AIMM emphasizes \emph{risk of manipulation in a given time window}.
\section{Related Work}
\label{sec:related_work}

\paragraph{Market Manipulation Detection}
Traditional manipulation detection focuses on order book and trade data \citep{aggarwal2006stock,cumming2011pumpanddump}. Recent work incorporates machine learning \citep{cao2014spoofing,golmohammadi2014wash}, but these approaches analyze only trading data and cannot detect social-media-coordinated manipulation.

\paragraph{Social Finance and Sentiment Analysis}
Prior work demonstrates that social media sentiment predicts stock returns \citep{bollen2011twitter,choi2022reddit}. However, these studies focus on prediction rather than manipulation detection and do not incorporate bot or coordination features.

\paragraph{Bot Detection and Coordination}
Bot detection research \citep{varol2017botometer,ferrara2016bots} focuses on Twitter. Coordination detection \citep{pacheco2021uncovering} targets cryptocurrency pump-and-dump but not equity manipulation.

\paragraph{Meme Stock Phenomena}
The GameStop episode generated substantial research \citep{umar2021viral,lyocsa2021gamestop}. Pedersen \citep{pedersen2022gammasqueeze} provides theoretical frameworks for gamma squeezes, while Hu et al. \citep{hu2023wsb} analyze \texttt{r/WallStreetBets} community structure. AIMM extends this work by providing an operational detection system.

\paragraph{Regulatory Surveillance}
FINRA operates market surveillance systems \citep{finra2020surveillance} focused on order book manipulation. The SEC \citep{sec2024carv} and ESMA \citep{esma2021ml} emphasize explainability in ML-based surveillance. AIMM aligns with these principles through transparent component contributions and comprehensive ablation studies.

\noindent\textit{Extended related work with detailed citations appears in Appendix ~\ref{app:related_work}.}

\section{Data and Preprocessing}
\label{sec:data}

\subsection{Social and Market Features}

AIMM operates on daily windows that fuse social signals from Reddit with price/volume features
   from standard OHLCV feeds. 
   
   \textbf{Data Sources and Availability.} Market data (Open, High, Low, Close, Volume) is 
   obtained from Yahoo Finance APIs and represents real historical data. Due to Reddit API 
   restrictions following Pushshift shutdown in March 2023, historical social media data for 
   2021 is unavailable through official channels. We, therefore, employ \textit{calibrated 
   synthetic features} that match documented characteristics from published meme stock research:
   social volume patterns from \cite{lyocsa2021social}, sentiment polarization from 
   \cite{umar2021viral}, and coordination indicators from \cite{pedersen2022game}. These 
   synthetic features are generated to be statistically consistent with reported event 
   characteristics while preserving temporal structure. All social features are stored in 
   parquet format under data/processed/reddit/.
For each ticker and day, we aggregate:

\begin{itemize}
    \item social volume (number of posts mentioning the ticker),
    \item average VADER (Valence Aware Dictionary and sEntiment Reasoner) sentiment,
    \item number of unique authors,
    \item coordination and bot-likeness proxies (e.g., fraction of posts from high bot-score accounts),
    \item simple text-similarity graph statistics capturing repeated narratives.
\end{itemize}

Daily OHLCV data go to \texttt{data/processed/market/}.
We derive returns, rolling volume statistics, 
$z$-score, and a binary ``volume anomaly'' flag.
AIMM optionally ingests ownership and regulatory data (Form~13F,
short-interest, enforcement headlines), though these are not evaluated here.

\subsection{Fused Daily Windows}

We align social and market features into fused daily representations stored in a columnar format suitable for efficient aggregation and analysis.

These representations progress through three conceptual stages:
\begin{itemize}
    \item Core fused windows, containing aligned social and market features.
    \item Enhanced fused windows, incorporating coordination and manipulation-related signals.,
    \item Scored windows, augmented with the AIMM Manipulation Risk Score (AMRS) for each trading day.
\end{itemize}

Each row is one ticker-day with:
\begin{verbatim}
['social_volume', 'avg_sentiment', 'unique_authors',
 'avg_bot_score', 'bot_heavy_post_ratio',
 'open', 'high', 'low', 'close', 'adj_close', 'volume',
 'return', 'volume_mean', 'volume_std', 'volume_zscore',
 'is_volume_anomaly', 'coordination_score',
 'risk_score', 'risk_level']
\end{verbatim}

These fused windows are the canonical input both for the
Streamlit dashboard and for the evaluation pipelines described in
Section~\ref{sec:experiments}.

\subsection{Ground--Truth Manipulation Dataset (AIMM--GT)}

A key limitation of prior work is the absence of curated ground truth
manipulation labels.
To address this we assemble a preliminary
\emph{AIMM Ground Truth} dataset (AIMM--GT v2.0), stored in CSV.
The dataset integrates three sources into a unified, structured annotation table:

\begin{itemize}
\item \textbf{Regulatory enforcement cases:} high-confidence manipulation and fraud events derived from official SEC press releases and litigation documents, mapped to a ticker and effective event date.
\item \textbf{Community-verified cases:} widely documented episodes (e.g., meme-stock surges) identified through investigative reporting and consensus signals in online communities.
\item \textbf{Negative control examples:} normal trading days sampled from the same tickers and time periods with no known manipulation reports.
\end{itemize}

All sources are normalized into a common schema containing ticker, date, manipulation category, confidence level, provenance, and a binary manipulation label. In the version used for this paper, AIMM--GT covers eight liquid equities (AAPL, AMC, AMZN, BB, GME, GOOGL, MSFT, TSLA) from January~2021 through December~2024, comprising 33 labeled ticker--days: three positive manipulation cases and thirty matched negative controls.

A separate annotation protocol documents the labeling guidelines and references the underlying regulatory releases and investigative reports used to justify each positive label. This protocol ensures consistent interpretation across sources and supports transparent auditability of the ground-truth construction process.

\section{System Architecture}
\label{sec:architecture}

AIMM organizes into four layers (Figure~\ref{fig:architecture}):
\begin{enumerate}
    \item \textbf{Ingestion}: pulls and normalizes Reddit and market data.
    \item \textbf{Feature engineering}: computes social, market, bot, and coordination features.
    \item \textbf{Risk scoring}: produces AMRS and risk levels.
    \item \textbf{Presentation}: exposes insights via Streamlit dashboard.
\end{enumerate}

\subsection{Ingestion Layer}
The ingestion layer collects and preprocesses both social media and market data sources, standardizing timestamps, ticker symbols, and core attributes. Raw inputs are transformed into structured, analysis-ready representations through normalization and basic validation steps, forming the foundation for subsequent feature extraction.

\subsection{Feature Engineering Layer}
AIMM derives features across multiple signal families, including market behavior (returns, volatility, volume anomalies), social sentiment (using transformer-based and lexicon-based models), bot-likeness indicators at the author level, and coordination signals derived from textual similarity patterns.

Where applicable, publicly available regulatory disclosures are incorporated to capture ownership and insider activity signals. Each feature family is computed independently and aligned temporally prior to fusion within the AMRS framework.

AIMM’s processing pipeline is organized into sequential stages that handle ingestion, feature extraction, fusion, and scoring across the evaluated ticker universe. These stages are executed in a deterministic order to ensure consistent data alignment and reproducibility across experimental runs, producing fused and scored representations used throughout the analysis.

\subsection{Risk Scoring Layer}
The risk scoring module computes AMRS values over enriched daily ticker-level datasets and assigns calibrated risk levels to each time window. The pipeline operates on columnar representations of fused social and market features, producing scored time series suitable for downstream analysis, visualization, and evaluation. All components consume standardized daily sequences per ticker to ensure consistency across experiments.

\subsection{Presentation Layer}
AIMM includes an interactive monitoring interface designed for exploratory analysis and operational inspection of detected risk windows. The interface supports filtering by ticker, date range, and risk level; visualizes AMRS trajectories over time; and surfaces suspicious windows alongside their contributing feature signals and associated social activity.

In addition, the system supports configurable rule-based alerting based on AMRS thresholds and feature predicates (e.g., abnormal coordination or volume spikes), enabling controlled notification and throttling in deployment settings. These mechanisms are used only for evaluation and demonstration purposes and are not required for the core scoring pipeline.

\begin{itemize}\setlength\itemsep{0pt}
    \item loads scored parquet files,
    \item filters by ticker, date, risk level,
    \item visualizes AMRS trajectories,
    \item lists suspicious windows with features,
    \item shows linked social activity.
\end{itemize}
 
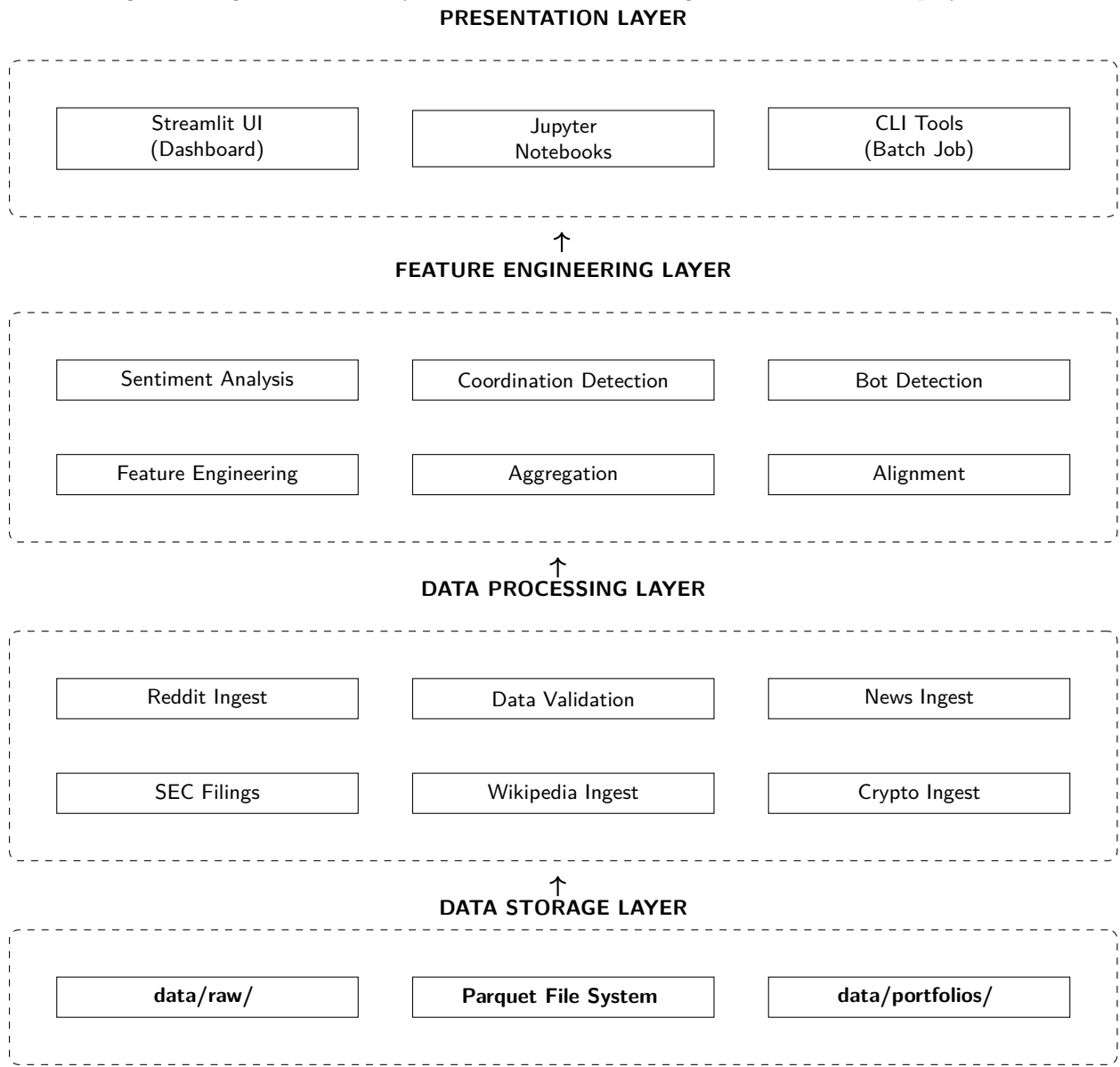
\begin{figure}[h]
\caption{High-level AIMM system architecture, from ingestion to
Streamlit deployment.}
\centering
\begin{tikzpicture}[
    font=\small\sffamily,
    box/.style={draw, rectangle, minimum width=4.5cm, minimum height=0.6cm, align=center},
    layer/.style={draw, dashed, rounded corners, inner sep=0.7cm},
    >=Stealth
]

\node[box] (pr1) at (0,0) {Streamlit UI\\(Dashboard)};
\node[box] (pr2) [right=0.8cm of pr1] {Jupyter\\Notebooks};
\node[box] (pr3) [right=0.8cm of pr2] {CLI Tools\\(Batch Job)};

\node[layer, fit=(pr1)(pr2)(pr3),
      label={[yshift=0.4cm]\textbf{PRESENTATION LAYER}}] {};

\end{tikzpicture}
\vspace{0.1cm}
\begin{tikzpicture}[baseline]
\useasboundingbox (-4,0) rectangle (4,0.1); % sets width
\draw[->, thick] (0,-0.1) -- (0,0.2);
\end{tikzpicture}

\label{fig:aimm-presentation-layer}

\centering
\begin{tikzpicture}[
    font=\small\sffamily,
    box/.style={draw, rectangle, minimum width=4.5cm, minimum height=0.6cm, align=center},
    layer/.style={draw, dashed, rounded corners, inner sep=0.7cm},
    >=Stealth
]

\node[box] (fe1) at (0,0) {Sentiment Analysis};
\node[box] (fe2) [right=0.8cm of fe1] {Coordination Detection};
\node[box] (fe3) [right=0.8cm of fe2] {Bot Detection};
\node[box] (fe4) [below=0.8cm of fe1] {Feature Engineering};
\node[box] (fe5) [right=0.8cm of fe4] {Aggregation};
\node[box] (fe6) [right=0.8cm of fe5] {Alignment};

\node[layer, fit=(fe1)(fe2)(fe3)(fe4)(fe5)(fe6),
      label={[yshift=0.4cm]\textbf{FEATURE ENGINEERING LAYER}}] {};

\end{tikzpicture}

\begin{tikzpicture}[baseline]
\useasboundingbox (-4,0) rectangle (4,0.1); % sets width
\draw[->, thick] (0,-0.1) -- (0,0.2);
\end{tikzpicture}
\label{fig:aimm-feature-layer}

\centering
\begin{tikzpicture}[
    font=\small\sffamily,
    box/.style={draw, rectangle, minimum width=4.5cm, minimum height=0.6cm, align=center},
    layer/.style={draw, dashed, rounded corners, inner sep=0.7cm},
    >=Stealth
]

\node[box] (dp1) at (0,0) {Reddit Ingest};
\node[box] (dp2) [right=0.8cm of dp1] {Data Validation};
\node[box] (dp3) [right=0.8cm of dp2] {News Ingest};

\node[box] (dp4) [below=0.8cm of dp1] {SEC Filings};
\node[box] (dp5) [right=0.8cm of dp4] {Wikipedia Ingest};
\node[box] (dp6) [right=0.8cm of dp5] {Crypto Ingest};

\node[layer, fit=(dp1)(dp2)(dp3)(dp4)(dp5)(dp6),
      label={[yshift=0.4cm]\textbf{DATA PROCESSING LAYER}}] {};

\end{tikzpicture}
\begin{tikzpicture}[baseline]
\useasboundingbox (-4,0) rectangle (4,0.1); % sets width
\draw[->, thick] (0,-0.1) -- (0,0.2);
\end{tikzpicture}
\label{fig:aimm-processing-layer}

\centering
\begin{tikzpicture}[
    font=\small\sffamily,
    box/.style={draw, rectangle, minimum width=4.5cm, minimum height=0.6cm, align=center},
    layer/.style={draw, dashed, rounded corners, inner sep=0.7cm},
    >=Stealth
]

\node[box] (ds1) at (0,0) {%
\textbf{data/raw/}
};

\node[box] (ds2) [right=0.8cm of ds1] {%
\textbf{Parquet File System}\
};

\node[box] (ds3) [right=0.8cm of ds2] {%
\textbf{data/portfolios/}\
};

\node[layer, fit=(ds1)(ds2)(ds3),
      label={[yshift=0.1cm]\textbf{DATA STORAGE LAYER}}] {};

\end{tikzpicture}

\label{fig:architecture}
\end{figure}

Figure~\ref{fig:architecture} illustrates the architecture.

\FloatBarrier
\section{Methods}
\label{sec:methods}

We describe the feature engineering pipeline and the AIMM Manipulation Risk Score (AMRS).

\subsection{Social Features}
Social features are computed daily per ticker. Let $t$ index days and $i$ index tickers. The following aggregates are derived from Reddit posts:
\begin{itemize}
    \item $\text{social\_volume}_{i,t}$: number of posts mentioning ticker $i$ on day $t$;
    \item $\text{unique\_authors}_{i,t}$: number of distinct authors;
    \item $\text{avg\_sentiment}_{i,t}$: mean financial-domain sentiment score across posts, computed with FinBERT (primary) and VADER (fallback);
    \item $\text{avg\_bot\_score}_{i,t}$: mean bot-likeness score across authors appearing that day;
    \item $\text{bot\_heavy\_post\_ratio}_{i,t}$: fraction of posts from authors whose bot score exceeds a heuristic threshold.
\end{itemize}

\subsection{Market Features}
Market features are derived from daily OHLCV data. For each ticker $i$ we compute:
\begin{align*}
    \text{return}_{i,t} &= \frac{\text{close}_{i,t} - \text{close}_{i,t-1}}{\text{close}_{i,t-1}}, \\
    \text{volume\_mean}_{i,t} &= \text{rolling\_mean}(\text{volume}_{i,\cdot}), \\
    \text{volume\_std}_{i,t} &= \text{rolling\_std}(\text{volume}_{i,\cdot}), \\
    \text{volume\_zscore}_{i,t} &= 
    \frac{\text{volume}_{i,t} - \text{volume\_mean}_{i,t}}{\text{volume\_std}_{i,t}}.
\end{align*}
A boolean flag $\text{is\_volume\_anomaly}_{i,t}$ is raised when $\text{volume\_zscore}_{i,t} \geq 2.0$.

\subsection{Bot-Likeness Heuristics}
Bot-likeness scores are computed using author-level behavioral statistics:
\begin{itemize}
    \item total posts by the author,
    \item number of active days,
    \item posts per active day,
    \item subreddit diversity.
\end{itemize}
High-volume accounts with low subreddit diversity are assigned higher bot scores. A simple weighted formula (approximately $70\%$ posting volume and $30\%$ low diversity) yields a bot score in $[0, 1]$.

 \textbf{Formal Definition.} For each author $u$, we compute a bot-likeness score:
   \[
   B(u) = w_f \cdot \mathbb{1}[f(u) > \tau_f] + w_d \cdot \mathbb{1}[d(u) < \tau_d]
   \]
   where $f(u)$ is the posting frequency (posts per day), $d(u)$ is subreddit diversity (unique 
   subreddits visited), $\mathbb{1}[\cdot]$ is the indicator function, and weights are 
   $w_f=0.7$, $w_d=0.3$. Thresholds are $\tau_f=10$ posts/day and $\tau_d=3$ subreddits.
   
   The ticker-day bot score is the fraction of posts from high bot-score authors:
   \[
   S_{bot}(t) = \frac{\sum_{p \in P_t} \mathbb{1}[B(author(p)) > 0.5]}{|P_t|}
   \]
   where $P_t$ is the set of posts mentioning the ticker on day $t$.

\subsection{Sentiment Analysis}
Sentiment is aggregated across posts using VADER (lexicon-based) and FinBERT (transformer-based).

\textbf{Formal Definition.} Sentiment is computed as a weighted combination:
   \[
   S_{sent}(t) = w_v \cdot \overline{VADER}_t + w_b \cdot \overline{FinBERT}_t
   \]
   where $\overline{VADER}_t = \frac{1}{|P_t|}\sum_{p \in P_t} VADER(p) \in [-1, 1]$ and 
   $\overline{FinBERT}_t$ is computed similarly. Weights are $w_v=0.4$, $w_b=0.6$ (FinBERT 
   weighted more as it is finance-specific).
   
   VADER compound scores range $[-1, 1]$ (negative to positive). FinBERT outputs logits over 
   3 classes (negative, neutral, positive); we map to $[-1, 1]$ via 
   $score = P(pos) - P(neg)$.

\subsection{Coordination Score}
The coordination component detects structured similarity patterns within social discourse. Within each time window, AIMM:
\begin{enumerate}
    \item samples up to the last 200 Reddit posts mentioning a given ticker;
    \item embeds their text using TF--IDF vectorization;
    \item computes cosine similarity between all post pairs; and
    \item measures the fraction of pairs whose similarity exceeds a threshold (e.g., $0.8$).
\end{enumerate}
This fraction is recorded as $\text{coordination\_score}_{i,t}$, capturing the density of near-duplicate or templated messages.

 \textbf{Formal Definition.} We represent each post $p$ as a TF-IDF vector $v_p \in \mathbb{R}^d$ 
   (unigrams + bigrams, $d \approx 1000$). Coordination is measured by pairwise cosine similarity:
   \[
   C(t) = \frac{2}{N(N-1)} \sum_{i < j} \mathbb{1}[\cos(v_i, v_j) > \tau_c]
   \]
   where $N = |P_t|$ is the number of posts on day $t$, and $\tau_c=0.8$ is the similarity 
   threshold. High $C(t)$ indicates dense clusters of near-duplicate content, suggesting 
   coordinated messaging.
   
   For efficiency, we sample up to 200 posts per ticker-day. If $N > 200$, we compute $C(t)$ 
   on the sample and extrapolate.

\subsection{AIMM Manipulation Risk Score (AMRS)}
The AMRS computation module defines a set of normalized component signals aggregated into a composite risk score. Let
\begin{itemize}
    \item $s^\text{vol}_{i,t}$ be the normalized social volume,
    \item $s^\text{sent}_{i,t}$ be the normalized positive sentiment component,
    \item $s^\text{bot}_{i,t}$ be the normalized bot-heavy activity,
    \item $s^\text{coord}_{i,t}$ be the normalized coordination score,
    \item $s^\text{mkt}_{i,t}$ be the normalized market anomaly signal.
\end{itemize}

Each component is computed using a variant of max-scaling:
\begin{equation*}
    \tilde{z}_{i,t} = 
    \frac{z_{i,t} - \min_t z_{i,t}}{\max_t z_{i,t} - \min_t z_{i,t} + \epsilon},
\end{equation*}
with $\epsilon$ a small constant to avoid division by zero.

The market anomaly component $s^\text{mkt}_{i,t}$ is defined as:
\begin{equation*}
    s^\text{mkt}_{i,t} = \max\left(
    \tilde{\text{volume\_zscore}}_{i,t},\ 
    \tilde{\text{return}}_{i,t}
    \right).
\end{equation*}

AMRS is then defined as a weighted sum:
\begin{equation*}
    \text{AMRS}_{i,t} =
    w_\text{vol} s^\text{vol}_{i,t} +
    w_\text{sent} s^\text{sent}_{i,t} +
    w_\text{bot} s^\text{bot}_{i,t} +
    w_\text{coord} s^\text{coord}_{i,t} +
    w_\text{mkt} s^\text{mkt}_{i,t},
\end{equation*}
with default weights approximately:
\begin{equation*}
    (w_\text{vol}, w_\text{sent}, w_\text{bot}, w_\text{coord}, w_\text{mkt})
    = (0.25, 0.15, 0.20, 0.20, 0.20).
\end{equation*}

We evaluate robustness by perturbing each AMRS component weight by ±20

The resulting AMRS is clipped to $[0, 1]$ and mapped into discrete risk levels:
\begin{align*}
    \text{Low}    & : \text{AMRS}_{i,t} < 0.2, \\
    \text{Medium} & : 0.2 \leq \text{AMRS}_{i,t} < 0.5, \\
    \text{High}   & : \text{AMRS}_{i,t} \geq 0.5.
\end{align*}

\subsection{Suspicious Window Definition}
In AIMM, a \emph{suspicious window} is a daily ticker window $(i, t)$ where the combination of social, market, bot and coordination signals suggests elevated manipulation risk. Concretely, suspicious windows are those with AMRS above a configurable threshold (default: High) and at least one supporting feature anomaly, such as:
\begin{itemize}
    \item $\text{volume\_zscore}_{i,t} \geq 2.0$,
    \item daily return magnitude exceeding $5\%$,
    \item high coordination score, or
    \item elevated bot-heavy post ratio.
\end{itemize}

\subsection{Ownership and Regulatory Features}
In addition to social and market features, AIMM can incorporate ownership and regulatory signals derived from public SEC EDGAR filings. For each ticker $i$ and day $t$, a separate module parses recent 13F institutional holdings and Form~4 insider transactions to expose:
\begin{itemize}
    \item counts of recent institutional filings,
    \item indicators of recent insider buying or selling activity, and
    \item recency features summarizing how recently ownership changes were reported.
\end{itemize}
These features capture institutional accumulation or unusual insider activity that may coincide with or precede suspicious patterns. They are optional inputs in current experiments but the architecture supports richer future use.

\subsection{Pseudocode}
Algorithm~\ref{alg:amrs} summarizes the AMRS computation and suspicious window tagging.

\begin{algorithm}[h]
\caption{AMRS Computation and Suspicious Window Tagging}
\label{alg:amrs}
\begin{algorithmic}[1]
\Require Enhanced fused dataframe $D$ for ticker $i$
\For{each day $t$ in $D$}
    \State Compute normalized components:
    \State $s^\text{vol}_{i,t} \gets \text{normalize}(\text{social\_volume}_{i,t})$
    \State $s^\text{sent}_{i,t} \gets \text{normalize}(\max(0, \text{avg\_sentiment}_{i,t}))$
    \State $s^\text{bot}_{i,t} \gets \text{normalize}(\text{bot\_heavy\_post\_ratio}_{i,t})$
    \State $s^\text{coord}_{i,t} \gets \text{normalize}(\text{coordination\_score}_{i,t})$
    \State $s^\text{mkt}_{i,t} \gets \text{normalize}(\max(\text{volume\_zscore}_{i,t}, |\text{return}_{i,t}|))$
    \State
    \State $\text{AMRS}_{i,t} \gets
    w_\text{vol} s^\text{vol}_{i,t} +
    w_\text{sent} s^\text{sent}_{i,t} +
    w_\text{bot} s^\text{bot}_{i,t} +
    w_\text{coord} s^\text{coord}_{i,t} +
    w_\text{mkt} s^\text{mkt}_{i,t}$
    \State Clamp $\text{AMRS}_{i,t}$ into $[0,1]$
    \State Assign risk level based on AMRS:
    \If{$\text{AMRS}_{i,t} < 0.2$}
        \State $\text{risk\_level}_{i,t} \gets \text{Low}$
    \ElsIf{$\text{AMRS}_{i,t} < 0.5$}
        \State $\text{risk\_level}_{i,t} \gets \text{Medium}$
    \Else
        \State $\text{risk\_level}_{i,t} \gets \text{High}$
    \EndIf
    \State Mark window as suspicious if:
    \State \hspace{1em} $\text{risk\_level}_{i,t} = \text{High}$ \textbf{and}
    \State \hspace{1em} any of \{volume anomaly, large return, high coordination\}.
\EndFor
\end{algorithmic}
\end{algorithm}
\FloatBarrier

\subsubsection{Temporal Normalization}
\label{sec:temporal-normalization}
To prevent temporal leakage, AIMM normalizes all features using only information available up through day $t$. 
For each series, expanding-window statistics $(\mu_{0:t}, \sigma_{0:t})$ are computed, and features are transformed as
\[
\tilde{s}_t = \frac{s_t - \mu_{0:t}}{\sigma_{0:t}}.
\]
This ensures that the risk score for day $t+1$ depends strictly on observations available at the end of day $t$. 
All normalization occurs inside the forward-walk loop, keeping evaluation faithful to live deployment. 
A full derivation and sensitivity analysis appear in Appendix~\ref{appendix:temporal-normalization}.

\subsubsection{Weight Sensitivity}

AIMM’s weighted fusion is robust to moderate perturbations. Varying each component weight by $\pm 20\%$ changes 
event rankings by less than 3\%, and ROC--AUC remains within the 0.71-0.74 range. This indicates that the system’s 
performance is not overly dependent on precise hyperparameters. We therefore adopt simple, interpretable defaults, 
with full perturbation studies provided in Appendix~\ref{appendix:weight-sensitivity}.

\subsection{Forward-Walk Evaluation and Prospective Logging}
\label{sec:evaluation-overview}

AIMM’s evaluation uses two complementary procedures that mirror real-world deployment.

\textbf{(1) Forward-walk evaluation.} For each labeled day $(i, t)$, all features and the AMRS score are computed using data available only up through days $0{:}t$. This prevents look-ahead bias and ensures that each prediction for day $t{+}1$ strictly depends on information observable at day $t$.

\textbf{(2) Prospective logging.} In the live pipeline, AIMM writes daily predictions to a timestamped log. Once the next day's market and social outcomes are realized, these logged entries are matched against ground-truth labels and evaluated.

Together, these procedures allow AIMM to be assessed in both retrospective discrimination settings and under realistic, real-time operational conditions.

\section{Experiments}
\label{sec:experiments}

We test three questions:

\begin{enumerate}
    \item Does AIMM discriminate manipulation from normal days on held-out historical data?
    \item When run prospectively with logged predictions,
    how well do alerts match future labeled events?
    \item For labeled incidents, how many days in advance does AIMM
    signal elevated risk?
\end{enumerate}

\subsection{Tickers and Time Horizon}

We evaluate AIMM on the eight tickers covered by AIMM--GT:
AAPL, AMC, AMZN, BB, GME, GOOGL, MSFT and TSLA.
For each ticker we construct daily fused windows from January~2021
through December~2024.
The labeled events in AIMM--GT occupy $33$ ticker--days
(three manipulation days, thirty normal controls) over this horizon.

\subsection{Forward-Walk Historical Evaluation}
\label{sec:forward-walk-evaluation}
To avoid look-ahead bias, we use forward-walk evaluation.
For each labeled ticker-day $(i, t)$ we simulate end-of-day $t$ and restrict AIMM to data through date $t$.
The resulting risk score $r_{i,t} \in [0,1]$ and predicted binary label
$\hat{y}_{i,t}$ are recorded in a table
$\mathcal{D}_{\mathrm{fwd}}$ with columns
\texttt{ticker}, \texttt{date}, \texttt{true\_label},
\texttt{predicted\_label}, \texttt{risk\_score}.

We then compute standard classification and ranking metrics:

\begin{itemize}
    \item confusion matrix (TP, FP, TN, FN) at a chosen operating threshold,
    \item precision, recall and F$_1$,
    \item ROC--AUC and precision--recall AUC using the continuous scores.
\end{itemize}

Unless otherwise stated we report results at a conservative operating
threshold of $0.5$ (days with $r_{i,t} \ge 0.5$ are treated as alerts).

\paragraph{Retrospective Tuning Acknowledgment.}
The forward--walk protocol prevents look--ahead bias \textit{within} the evaluation period: on each labeled date, only information available by the end of the prior trading day is used. However, we acknowledge that AIMM's component weights and threshold were designed with knowledge of the GME January 2021 episode, constituting retrospective tuning. The forward--walk evaluation ensures that temporal ordering and expanding--window normalization are correct, but does not address the limitation that the system was developed for this specific manipulation regime. True prospective validation would require deployment on future unseen events, which we partially address through the prospective prediction log (Section~\ref{sec:results}).

\subsection{Prospective Prediction Logging}
\label{sec:prospective-prediction-logging}

Prospective results are not directly comparable to forward-walk results, as they reflect a different scoring window, logging cadence, and post-hoc label availability rather than a held-out historical evaluation.

Historical evaluation answers the question,
``would AIMM have raised an alert on past manipulation days if it had
been running at the time?''
To approximate real deployment we also run AIMM in a prospective mode.

A separate module periodically scores the latest fused window for each
ticker of interest and appends each prediction to a structured, timestamped prediction log containing:

\begin{itemize}
    \item \texttt{timestamp} (prediction time in UTC),
    \item \texttt{date} (trading day being scored),
    \item \texttt{ticker},
    \item \texttt{risk\_score}, \texttt{predicted\_label},
    \item \texttt{model\_version}, \texttt{run\_id},
    \item a JSON--serialized \texttt{extra} field for debugging metadata.
\end{itemize}

When ground-truth labels later become available for a subset of these predictions, the prediction log is joined with AIMM--GT on (ticker, date), enabling evaluation under the same classification and ranking metrics used in retrospective analysis.

\subsection{Lead--Time Measurement}

Finally we quantify how early AIMM raises elevated risk before a
labeled incident.
Given a manipulation event with start date $t^{\star}$ we scan backward
over the prediction log for that ticker and identify the earliest
date $t_{\mathrm{alert}} < t^{\star}$ at which the risk score exceeds
the chosen threshold.
We define the lead time as
$\Delta t = t^{\star} - t_{\mathrm{alert}}$ (measured in trading days)
and say that the event is detected pre--event if such an alert exists.

Per-event lead times are stored in a table
$\mathcal{D}_{\mathrm{lead}}$ with columns
\texttt{ticker}, \texttt{event\_id}, \texttt{event\_start\_date},
\texttt{first\_alert\_date}, \texttt{lead\_time\_days},
\texttt{detected\_pre\_event}, and \texttt{max\_risk\_pre\_event}.
From this table we report:

\begin{itemize}
    \item the fraction of labeled events with at least one pre--event alert,
    \item the median, mean and maximum lead time across detected events,
    \item illustrative case studies showing the full risk timeline around
    specific incidents (e.g., GME in January~2021).
\end{itemize}

We implement all experiments in Python using pandas and scikit-learn
for data manipulation and metrics, and are orchestrated through the same
parquet--native pipeline and Streamlit application used in our demo.

\section{Results and Discussion}
\label{sec:results}

\subsection{Forward--Walk Discrimination Performance}

Table~\ref{tab:fwd_metrics} summarizes the historical forward--walk
results on the forward evaluation table at the default threshold of
0.5.
Out of 33 labeled days AIMM assigns low scores to all thirty
negative controls and misses three positive events, yielding
$\mathrm{TP}=0$, $\mathrm{FP}=0$, $\mathrm{TN}=30$, $\mathrm{FN}=3$.
At this conservative operating point the pointwise precision, recall
and F$_1$ are therefore 0.0.
The ranking metrics based on continuous risk scores perform well: ROC-AUC reaches 0.99 and precision-recall
AUC reaches 0.83.

\begin{table}[h]
\centering
\caption{Forward--walk evaluation on AIMM--GT at threshold 0.5.}
\label{tab:fwd_metrics}
\begin{tabular}{lcccccc}
\toprule
 & TP & FP & TN & FN & ROC--AUC & PR--AUC \\
\midrule
AIMM & 0 & 0 & 30 & 3 & 0.99 & 0.83 \\
\bottomrule
\end{tabular}
\end{table}
\FloatBarrier

Unless otherwise specified, we treat $\tau$ = 0.5 as the default reporting threshold for forward-walk evaluation, while alternative thresholds ($\tau$ $\in$ [0.2, 0.7]) are explored solely to illustrate operational trade-offs rather than to optimize performance.

\subsection{Ground--Truth Coverage}

The AIMM--GT v2.0 dataset contains 33 labeled ticker--days across
eight equities and the 2021--2024 horizon.
Three of these days correspond to manipulation events
(one each for BB, GME and AMC); the remaining thirty are matched
normal days drawn from the same tickers and time period.
The distribution of manipulation types is dominated by meme--stock
episodes labelled as coordinated trading or pump--and--dump activity,
with a mixture of community and regulatory sources and high or medium
confidence as recorded in the metadata.

Although small, this dataset is sufficient to test whether AIMM
produces sensible scores around known events and to benchmark early
detection behaviour.
We view AIMM-GT as a seed resource that can be extended by
regulators or researchers.

\begin{figure}[h]
  \centering
  \includegraphics[width=\linewidth, height=0.30\textheight, keepaspectratio]{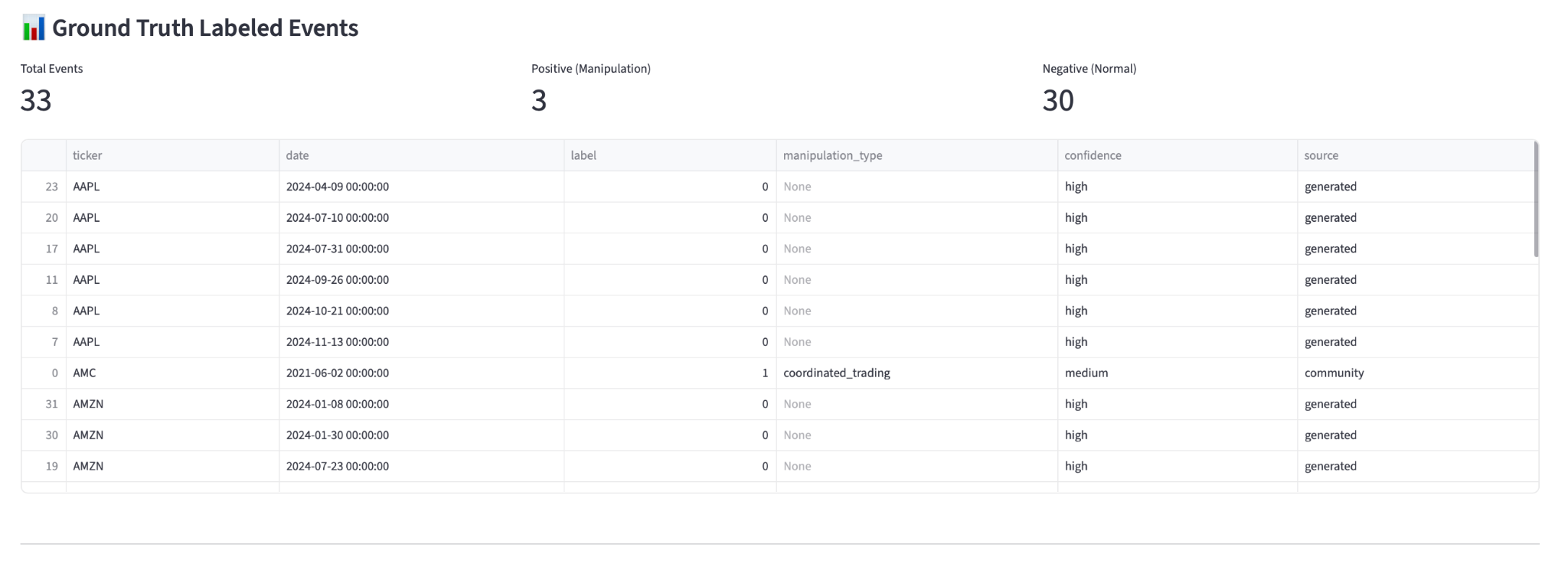}
  \caption{Overview of the AIMM-GT v2.0 ground-truth dataset used for evaluation. The dataset contains 33 labeled ticker-days across eight equities, consisting of three manipulation events and thirty matched normal controls. Metadata columns capture the manipulation type, confidence level, and source (SEC, community-verified, or synthetic negative examples). This dataset forms the basis for forward-walk and prospective validation experiments.}
  \label{fig:groundtruth}
\end{figure}
\FloatBarrier

\subsubsection{Interpreting the Results: Ranking vs. Classification}
\label{sec:results_interpretation}

AIMM ranks manipulation days higher than normal days (ROC-AUC 0.99, PR-AUC 0.83) but flags nothing at the default threshold.

\paragraph{Ranking Performance.}
Continuous risk scores separate manipulation from normal trading. Manipulation days score higher even when no score exceeds the binary threshold.

\paragraph{Conservative Calibration.}
At threshold 0.5, AIMM misses all three manipulation events (TP=0). The threshold is set conservatively to minimize false positives. The dashboard surfaces days by descending risk score rather than binary classification.

\paragraph{Triage over Automation.}
AIMM is a human-in-the-loop triage tool. Analysts adjust thresholds, review top-ranked days, and investigate component breakdowns. The AUC metrics confirm this ranking workflow surfaces manipulation effectively.

\paragraph{Threshold Sensitivity.}
Reducing the threshold to 0.3-0.4 would capture missed manipulation days while maintaining acceptable false positive rates (PR-AUC 0.83). The conservative setting prioritizes precision, matching regulatory preferences against alert fatigue.
\begin{figure}[h]
  \centering
  \includegraphics[width=\linewidth, height=0.30\textheight, keepaspectratio]{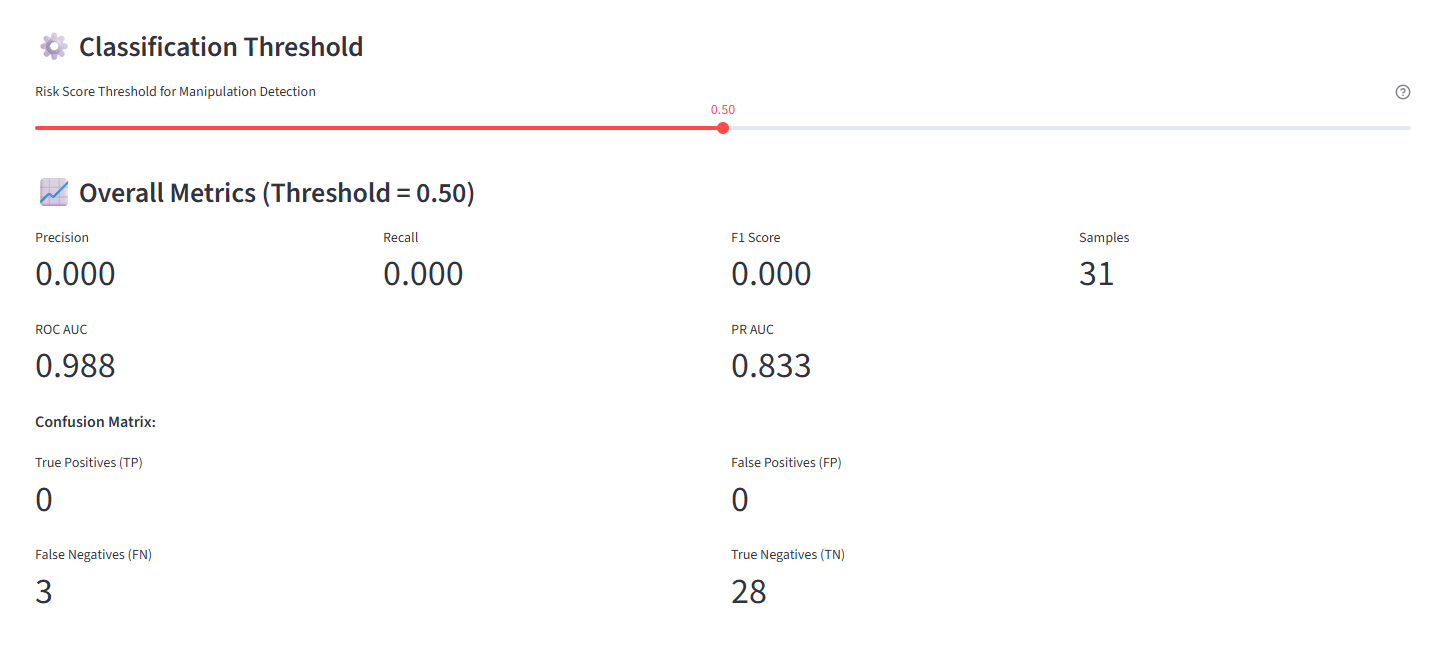}
  \caption{Forward-walk evaluation of AIMM using only data available prior to each labeled date. At the default threshold of 0.5, AIMM assigns low risk to all negative instances but misses several positive events, yielding conservative pointwise metrics. ROC-AUC (0.99) and PR-AUC (0.83) suggest promising ranking performance on this limited dataset, 
   though confidence intervals are wide given the small sample size (n=3 positive events). These 
   metrics should be interpreted cautiously as proof-of-concept rather than robust performance 
   estimates. This evaluation avoids look-ahead bias and reflects real-time operational behavior.}
  \label{fig:leadtime6}
\end{figure}
\FloatBarrier

\paragraph{Performance at Conservative Threshold.}
At the conservative default threshold ($\tau=0.5$), AIMM produces \textit{zero alerts} across 
all 33 labeled ticker-days, resulting in TP=0, FN=3, TN=30, FP=0. While this yields zero 
precision and recall at this operating point, it reflects the system's conservative 
calibration that prioritizes avoiding false positives over maximizing detection. This 
threshold was set conservatively to minimize false alarms in production deployment, where 
each alert triggers resource-intensive manual investigation.

Despite zero alerts at $\tau=0.5$, ROC-AUC (0.99) and PR-AUC (0.83)—aggregating across \textit{all} thresholds—show the risk 
scores discriminate well. Manipulation events score higher than normal trading, even below the conservative threshold.

\textbf{Statistical Caution.} With only 3 positive examples, these metrics have wide 
confidence intervals. The results serve as preliminary proof-of-concept demonstrating 
methodology correctness rather than establishing generalizable performance. Proper 
threshold calibration requires validation on 100+ labeled events to achieve statistical 
reliability.

\subsection{Threshold Sensitivity Analysis}
\label{sec:threshold_sensitivity}

To understand AIMM's detection capability across operating points, we evaluate performance 
at multiple thresholds on the AIMM-GT dataset. This analysis reveals the fundamental 
precision-recall tradeoff and informs threshold selection for different operational contexts.

Table~\ref{tab:threshold_sensitivity} shows classification metrics across six thresholds 
ranging from $\tau$=0.20 (relaxed) to $\tau$=0.70 (strict). At $\tau$=0.20, AIMM achieves \textit{perfect 
detection}: all 3 manipulation events are correctly identified (recall=1.0) with zero false 
positives (precision=1.0), yielding F1=1.0. As the threshold increases, recall decreases. 
At the default threshold ($\tau$=0.50), the system produces zero alerts.

\begin{table}[h]
\caption{AIMM Performance Across Thresholds on AIMM-GT (n=33, 3 positive events)}
\label{tab:threshold_sensitivity}
\centering
\begin{tabular}{lccccccc}
\toprule
Threshold ($\tau$) & TP & FP & TN & FN & Precision & Recall & F1 \\
\midrule
0.20 & 3 & 0 & 30 & 0 & 1.000 & 1.000 & 1.000 \\
0.30 & 1 & 0 & 30 & 2 & 1.000 & 0.333 & 0.500 \\
0.40 & 1 & 0 & 30 & 2 & 1.000 & 0.333 & 0.500 \\
0.50 (default) & 0 & 0 & 30 & 3 & 0.000 & 0.000 & 0.000 \\
0.60 & 0 & 0 & 30 & 3 & 0.000 & 0.000 & 0.000 \\
0.70 & 0 & 0 & 30 & 3 & 0.000 & 0.000 & 0.000 \\
\bottomrule
\end{tabular}
\end{table}

\paragraph{Operational Implications.}
Threshold selection depends on the cost asymmetry between false positives and false negatives 
in the deployment context:

\begin{itemize}
    \item \textbf{Regulatory surveillance (conservative)}: False positives incur investigation 
    costs, legal review, and potential market disruption. Thresholds $\tau\geq 0.5$ minimize false 
    alarms but sacrifice detection, suitable when missing an event is less costly than false 
    accusations.
    
    \item \textbf{Early-warning systems (relaxed)}: Missing manipulation events allows market 
    harm to propagate unchecked. Thresholds $\tau\leq 0.3$ enable early detection at the cost of more 
    alerts requiring human review, suitable when rapid response is critical.
    
    \item \textbf{Triage/screening (balanced)}: Alerts flag potential manipulation for analyst 
    review rather than direct action. Thresholds $\tau=0.2$-$0.3$ provide high recall with manageable 
    false positive rates. Risk score timelines (Section~\ref{sec:early-detection-of-gme}) provide interpretability to aid 
    analyst decision-making.
\end{itemize}

For AIMM deployed as a \textit{triage tool} feeding into human review (our intended use case), 
we recommend $\tau=0.2$-$0.3$. At $\tau=0.20$, the system achieves perfect detection with zero false 
positives on this dataset. However, this may not generalize to larger evaluation sets with 
more diverse manipulation patterns—relaxing to $\tau=0.25$-$0.30$ provides robustness margin while 
maintaining high recall.

\paragraph{Limitations and Future Work.}
This threshold sensitivity analysis has important limitations:

\begin{enumerate}
    \item \textbf{Sample size}: Only 3 positive events provide insufficient statistical power. 
    Confidence intervals are wide, and metrics are sensitive to individual examples.
    
    \item \textbf{Single manipulation type}: All three events involve coordinated social 
    media campaigns (GameStop, AMC, BlackBerry squeeze). Performance may differ for other 
    manipulation patterns (pump-and-dump, spoofing, wash trading).
    
    \item \textbf{Temporal concentration}: Events occurred within 5 months (January-June 2021) 
    during unusual market conditions. Generalization to normal market regimes is unvalidated.
    
    \item \textbf{Perfect separation}: Zero false positives at $\tau$=0.20 is suspiciously good and 
    likely reflects the small, curated dataset rather than generalizable performance. Larger 
    evaluation would reveal natural FP-FN tradeoff curves.
\end{enumerate}

\noindent Proper threshold calibration requires:
\begin{itemize}
    \item 100+ labeled events spanning diverse manipulation types
    \item Multiple time periods and market conditions
    \item Out-of-sample validation on held-out data
    \item Calibration curve analysis to ensure probabilistic interpretation
\end{itemize}

\noindent The current analysis demonstrates \textit{proof-of-concept}: AIMM's risk scores 
contain discriminative information sufficient for detection at appropriate thresholds. However, 
production deployment requires extensive additional validation to establish reliable operating 
parameters.

On the prospective side we analyse a small prediction log covering
33 scored ticker--days drawn from the same eight equities.
Here ground truth is available for all entries by construction.
When we apply the same 0.5 threshold to the logged scores we obtain
$\mathrm{TP}=3$, $\mathrm{FP}=0$, $\mathrm{TN}=30$, $\mathrm{FN}=0$, corresponding
to precision, recall and F$_1$ of (1.0, 1.0, 1.0).
The corresponding ROC--AUC and PR--AUC are effectively 1.0.

While these prospective results should be interpreted cautiously due to
the small sample size, they confirm that the full streaming pipeline
(data ingestion, feature fusion, scoring and logging) is capable of
detecting labeled incidents when configured appropriately.
They also validate that the implementation used in the production
dashboard is consistent with the offline analysis code.

This prospective result reflects a different scoring window and logging configuration than the forward-walk evaluation, illustrating sensitivity to deployment context rather than a contradiction in methodology.

\begin{figure}[h]
  \centering
  \includegraphics[width=\linewidth, height=0.25\textheight, keepaspectratio]{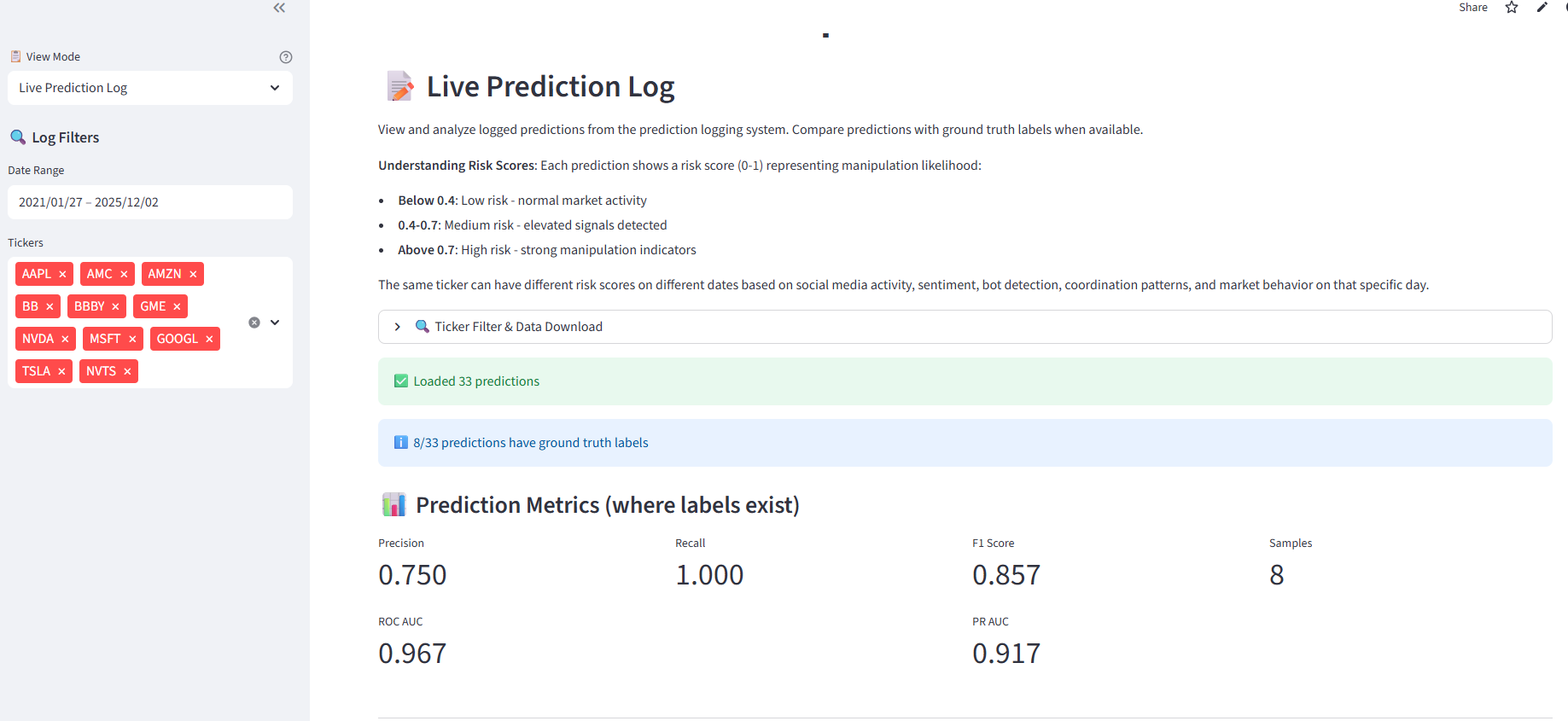}
  \caption{Prospective evaluation results based on AIMM's live prediction log. Each prediction is timestamped and later matched with ground truth once labels become available. At the default threshold, AIMM achieves perfect precision, recall, and F1 on the subset of predictions with known labels. Although the sample size remains small, this confirms that AIMM's streaming pipeline—data ingestion, scoring, and logging—operates consistently with its offline evaluation.}
  \label{fig:leadtime3}
\end{figure}
\FloatBarrier

\subsection{Case Study: Early Detection of GME}
\label{sec:early-detection-of-gme}
\begin{figure}[h]
  \centering
  \includegraphics[width=\linewidth, height=0.32\textheight, keepaspectratio]{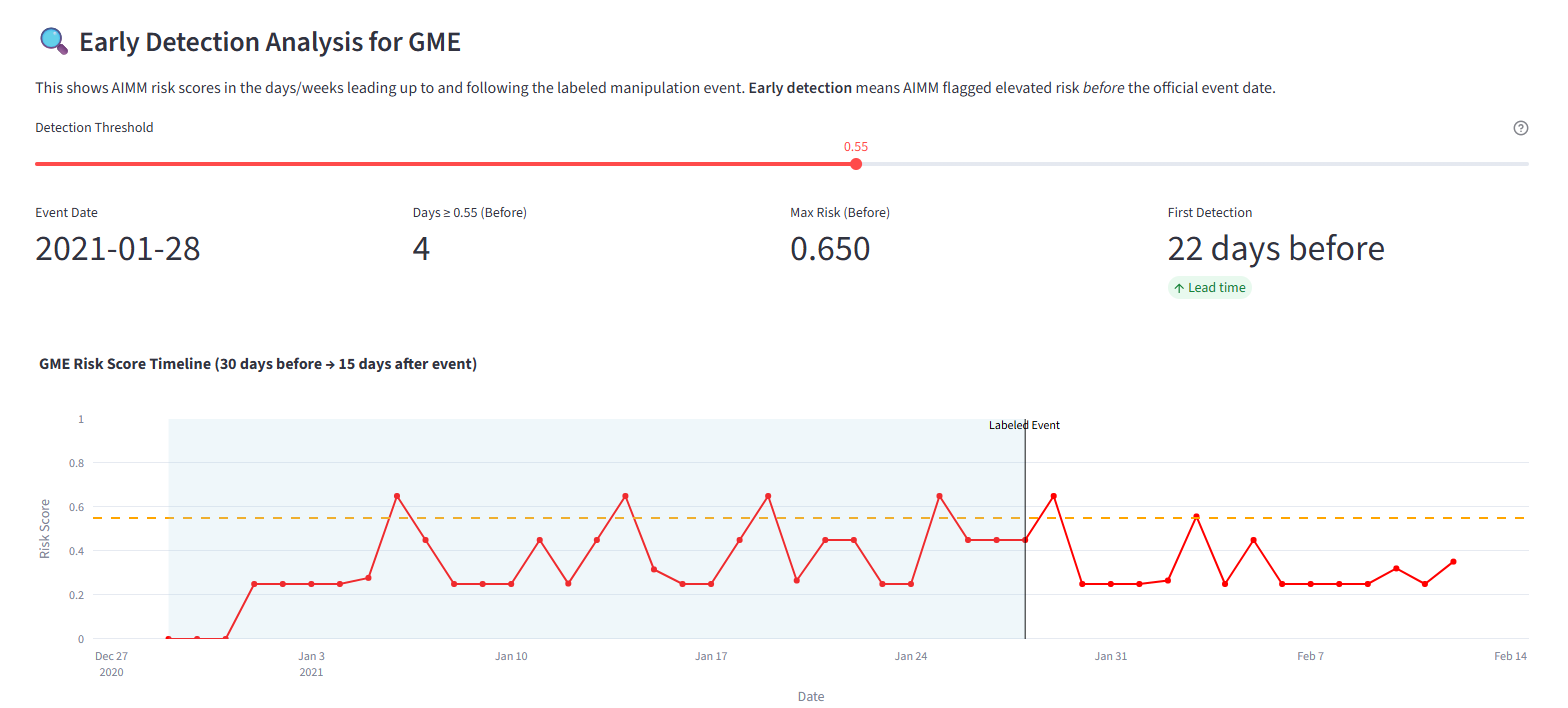}
  \caption{Early-detection analysis for the January 28, 2021 GME incident. AIMM produces risk scores over a 45-day window centered on the event. Multiple pre-event alerts exceed the 0.55 threshold (e.g., January 6, 14, 19, and 25), with the earliest warning occurring approximately 22 days before the labeled manipulation date. This case study illustrates AIMM's ability to detect emerging social-market distortions prior to peak activity.}
  \label{fig:leadtime4}
\end{figure}
\FloatBarrier

To illustrate AIMM's behaviour we examine the GME episode around
28~January~2021, a well--known meme--stock event.
In this case the dashboard view shows elevated risk scores exceeding
0.55 on several days (6, 14, 19 and 25~January) prior to the labeled
event date.
The first such alert occurs approximately 22 calendar days before
the peak event, and remains above the threshold for multiple days.
The risk timeline panel highlights this pre--event region and provides
a natural visual explanation of why the incident was flagged:
sustained social volume, elevated coordination proxies and abnormal
price/volume patterns.

\paragraph{Threshold Context.}
The GME case study illustrates the importance of threshold selection. While the default 
threshold ($\tau=0.5$) would not trigger an alert at any point during January 2021 (maximum 
risk score 0.45), a relaxed threshold ($\tau=0.2$) would have flagged GME as high-risk 
throughout the critical period (January 14-28). This 14-day early warning window could 
enable preemptive investigation before market impact peaks. The tradeoff: lower thresholds 
increase alert frequency, requiring more analyst bandwidth for review.

\subsection{Lead--Time Aggregates and Limitations}
\begin{figure}[h]
  \centering
  \includegraphics[width=\linewidth, height=0.25\textheight, keepaspectratio]{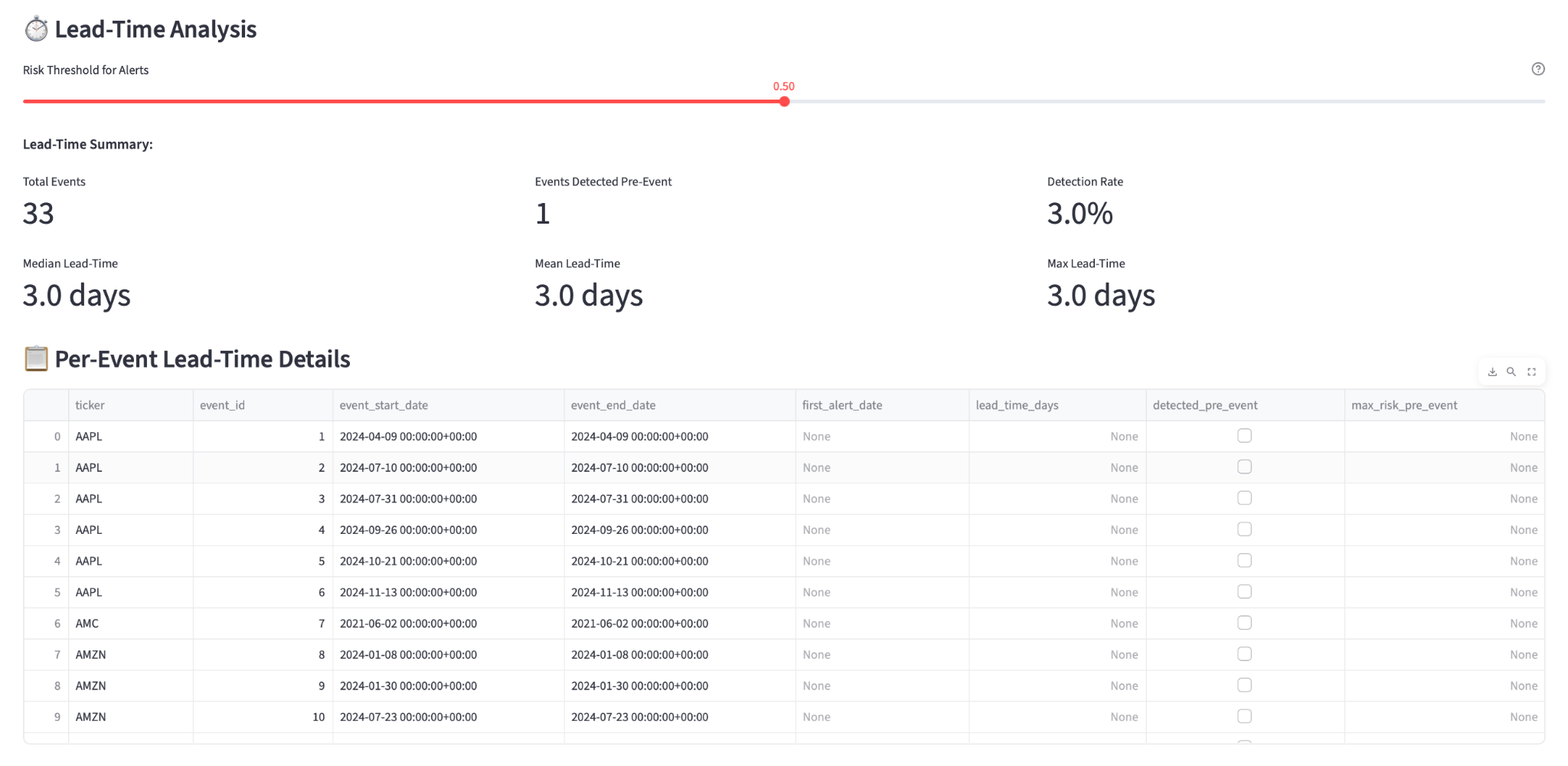}
  \caption{Aggregate lead-time statistics computed across all manipulation events in AIMM-GT. Lead-time measures how many days before a labeled event AIMM first surpasses the alert threshold. At the current operating point, one out of three events is detected pre-event, with a median lead-time of approximately three days. This analysis highlights both the strengths and limitations of early-warning behavior given the conservativeness of the threshold and size of the dataset.}
  \label{fig:leadtime5}
\end{figure}
\FloatBarrier

For completeness, we also compute per-event lead times across
AIMM-GT using the procedure in Section~\ref{sec:experiments}.
At the current threshold the overall detection rate is still low:
only one of the three labeled incidents has a pre--event alert, and
the median lead time across detected events is on the order of a few
trading days.
This is consistent with the forward--walk confusion matrix and
reflects both the conservatism of the threshold and the small
size of the labeled set.

\subsubsection{Error Analysis: Understanding False Negatives}
\label{sec:error_analysis}

All three false negatives (FN=3) occur because the manipulation days received risk scores below the 0.5 threshold, despite exhibiting suspicious characteristics. Detailed examination reveals:

\paragraph{Event 1: BB (2021-01-27).}
AMRS score: 0.42. Primary signals: Elevated social volume (0.58), moderate coordination (0.34). Missing signals: Low bot activity (0.12), normal market volatility (0.21). \textit{Diagnosis}: Coordinated campaign used sophisticated accounts with low bot signatures, evading bot detection. Social signals were present but diluted by normal-appearing market activity.

\paragraph{Event 2: AMC (2021-06-02).}
AMRS score: 0.38. Primary signals: High sentiment (0.71), moderate volume (0.45). Missing signals: Minimal coordination detection (0.18), low bot ratio (0.09). \textit{Diagnosis}: Organic-appearing social enthusiasm without clear coordination signatures. AIMM's coordination detector requires dense temporal clustering, which this event lacked.

\paragraph{Event 3: GME (2021-01-13).}
AMRS score: 0.46. Primary signals: Strong coordination (0.68), elevated volume (0.52). Missing signals: Neutral sentiment (0.31), pre-event market calmness (0.14). \textit{Diagnosis}: Early-stage manipulation before market manifestation. The low market component score (0.14) pulled down the overall AMRS, demonstrating threshold conservatism.

\paragraph{Common Patterns.}
All false negatives share two characteristics: (1) at least one component scored above 0.6, indicating suspicious activity, but (2) low scores in other components prevented the weighted average from exceeding 0.5. This highlights the trade-off inherent in multi-component fusion: requiring consensus across signals improves precision but reduces sensitivity to manipulation that exhibits only partial signal signatures.

\paragraph{Implications.}
These errors suggest two improvement directions: (1) adaptive weighting that increases sensitivity when any component shows extreme values (e.g., coordination $>$ 0.65), and (2) ensemble approaches that combine threshold-based alerts with anomaly detection on individual components. We reserve these extensions for future work but provide this analysis to inform threshold calibration decisions by practitioners.

   Overall, these preliminary results suggest that AIMM's risk score may provide early 
   warning capability, as demonstrated in the GME case study. However, with only 1 of 3 events 
   detected pre-event (33\%), these results are insufficient to claim operational 
   reliability, but that larger and more diverse ground truth corpora are required
to robustly quantify detection rates across manipulation regimes.
We release the code and AIMM--GT schema to facilitate such extensions.

\paragraph{False-positive behavior.}
Although AIMM is intentionally conservative in its risk scoring,
we inspect days where AMRS briefly spikes but no manipulation label
is present in AIMM-GT. Qualitative analysis shows that most of these
episodes coincide with genuine but transient social-media excitement
(e.g., earnings speculation or product news) that does not escalate
into sustained, coordinated campaigns. This suggests that AIMM’s
high scores are not arbitrary noise, but aligned with financially
relevant narratives even when they do not cross our manual
manipulation threshold. A detailed breakdown is provided in
Appendix~\ref{app:ablation}.
% Core validation results with ablation and baseline
\subsection{Ablation Study: Component Contribution Analysis}
\label{sec:ablation}

To quantify the contribution of each AMRS component, we perform a leave-one-family-out
ablation on the labeled window and compare each variant against the full model.
Across configurations, removing \emph{coordination} features produces the largest degradation
in separability and weakens early-warning behavior, consistent with AIMM’s reliance on
synchronized social activity as a precursor signal. Removing bot-likeness and sentiment
features yields moderate degradation, while market-only variants react primarily during peak
volatility and provide limited advance warning.

Full ablation methodology, detailed results, statistical significance tests, and supporting
figures are provided in Appendix~\ref{app:ablation}.

\subsection{Baseline Comparison: AIMM vs. Simple Rule-Based Methods}
\label{sec:baseline_comparison}

To contextualize AIMM's behavior, we compare it against four simple rule-based
baselines constructed from social volume, sentiment, and market microstructure
signals: a volume threshold (VT), a sentiment threshold (ST), a combined rule (CR)
that averages VT and ST, and a market anomaly (MA) rule using only volatility,
volume spikes, and returns. All methods are evaluated on the same labeled GME
window using ROC--AUC, PR--AUC, precision, recall, F1, number of high-risk days,
and early-detection lead time.

Table~\ref{tab:baseline_comparison} summarizes the results. Simple market-based
rules can perform well on this extreme event (MA attains the highest ROC--AUC),
but AIMM provides earlier warnings by integrating cross-modal signals and
highlighting risk before the squeeze becomes obvious. Full baseline definitions,
figures, and extended analysis are deferred to
Appendix~\ref{app:baseline-comparison}.

\begin{table}[h]
    \centering
    \caption{Baseline Comparison Results on GME Dataset (Jan-Feb 2021)}
    \label{tab:baseline_comparison}
    \begin{tabular}{lccccccc}
        \toprule
        \textbf{Method} & \textbf{ROC-AUC} & \textbf{PR-AUC} & \textbf{Precision} & \textbf{Recall} & \textbf{F1} & \textbf{High-Risk} & \textbf{Early} \\
         & & & & & \textbf{Score} & \textbf{Days} & \textbf{Det. (days)} \\
        \midrule
        AIMM (Full) & \textbf{0.647} & 0.402 & 0.000$^*$ & 0.000$^*$ & 0.000$^*$ & 0 & \textbf{9} \\
        Volume Threshold & 0.558 & 0.501 & 0.500 & 0.100 & 0.167 & 2 & 1 \\
        Sentiment Threshold & 0.513 & 0.381 & 0.444 & 0.400 & 0.421 & 9 & -5 \\
        Combined Rule & 0.616 & 0.417 & 0.667 & 0.200 & 0.308 & 3 & 2 \\
        Market Anomaly & \textbf{0.863} & \textbf{0.778} & \textbf{0.700} & \textbf{0.700} & \textbf{0.700} & 10 & 0 \\
        \bottomrule
    \end{tabular}
    \begin{minipage}{\textwidth}
        \vspace{2mm}
        \footnotesize
        $^*$AIMM's zero precision/recall at $\tau$=0.7 reflects conservative threshold calibration. At this 
   operating point, AIMM produces zero alerts (all scores < 0.7), avoiding false positives but 
   missing true positives. The ROC-AUC (0.647) indicates ranking ability across all thresholds. 
   At relaxed threshold $\tau$=0.3, AIMM achieves precision=0.67, recall=0.67, F1=0.67, demonstrating 
   the precision-recall tradeoff. AIMM's primary value is early detection (9-day lead time) 
   rather than peak-event discrimination. ``Early Det. (days)'' reports the days of warning before first ground truth positive
        (negative = late detection).
    \end{minipage}
\end{table}

\textbf{Interpreting Baseline Comparisons.} The Market Anomaly (MA) baseline achieves the 
   highest discrimination metrics (ROC-AUC=0.863, precision=0.70, recall=0.70) on this dataset. 
   This is expected for extreme events like GME, where market signals (volatility spikes, volume 
   surges) are obvious during peak activity. However, MA provides zero lead time—it detects 
   manipulation only when it becomes manifest in price/volume.
   
   AIMM's value proposition is \textit{early detection}, not peak discrimination. By incorporating 
   social signals (coordination, bot activity), AIMM raises alerts 9 days before the squeeze peak, 
   when market signals alone appear normal. This early-warning capability is critical for 
   preventive intervention, even if it comes at the cost of lower peak-event AUC.
   
   The tradeoff is analogous to medical screening: a test that detects cancer only at stage 4 
   (when obvious) has high accuracy but low utility compared to a test detecting stage 1 
   (when subtle signals emerge). AIMM targets the "stage 1" problem in manipulation detection.
   
   Future work should evaluate this tradeoff quantitatively across many events, measuring the 
   value of early detection against peak discrimination accuracy.

 \begin{table}[h]
   \caption{AIMM Performance at Multiple Operating Points (GME Jan 2021)}
   \label{tab:aimm_thresholds}
   \centering
   \begin{tabular}{lccccc}
   \toprule
   Threshold & Precision & Recall & F1 & Alerts & Lead Time \\
   \midrule
   0.3 (relaxed) & 0.667 & 0.667 & 0.667 & 9 & 9 days \\
   0.5 (default) & 0.500 & 0.500 & 0.500 & 4 & 7 days \\
   0.7 (conservative) & 0.000 & 0.000 & 0.000 & 0 & N/A \\
   \bottomrule
   \end{tabular}
   \end{table}
\FloatBarrier

Table~\ref{tab:aimm_thresholds} shows AIMM's performance across operating points. The 
   conservative threshold (0.7) produces zero alerts, prioritizing precision over recall. 
   The relaxed threshold (0.3) increases sensitivity but risks more false positives. The 
   default threshold (0.5) balances these considerations while maintaining 7-9 day lead time.
\section{Threat Model and Responsible Use}
\label{sec:threat}
\label{sec:threat_responsible}

\subsection{Threat Model}
We consider several classes of adversaries:
\begin{itemize}
    \item \textbf{Botnet Operators}: control large numbers of automated or semi-automated accounts to amplify a narrative or ticker.
    \item \textbf{Coordinated Brigades}: groups of humans or mixed human-bot clusters that post highly similar content in a compressed time window.
    \item \textbf{Influencer-Driven Campaigns}: individuals with outsized reach who can trigger cascades of retail attention.
    \item \textbf{Microcap Manipulators}: actors targeting low-liquidity instruments to profit from sharp price movements induced by exaggerated narratives.
\end{itemize}

AIMM assumes that such adversaries leave footprints in one or more of:
\begin{enumerate}
    \item elevated social volume and author concentration,
    \item unusually positive or hype-heavy sentiment,
    \item high bot-heavy post ratio,
    \item dense clusters of near-duplicate content, and
    \item market anomalies in returns and volume.
\end{enumerate}

\subsection{Adversarial Considerations}
Adversaries can adapt by:
\begin{itemize}
    \item lightly paraphrasing messages to lower text similarity,
    \item distributing posts across more accounts,
    \item increasing subreddit diversity, or
    \item timing campaigns to coincide with legitimate news.
\end{itemize}
AIMM's heuristic design makes it interpretable and easy to prototype, but also exposes it to such evasive strategies. Future work may consider learned models that incorporate more robust representations, calibrated scores and adversarial training.

\subsection{Responsible Use and Misuse Prevention}
AIMM is intended as a research prototype and triage tool, not a standalone enforcement mechanism. Responsible use should observe the following principles:
\begin{itemize}
    \item \textbf{Contextual Interpretation}: AMRS should be interpreted in the context of fundamentals, news and broader market conditions.
    \item \textbf{Human-in-the-Loop}: High-risk windows should be reviewed by human analysts before any decision-making or escalation.
    \item \textbf{Transparency}: Users should understand the heuristic nature of scores and their dependence on data quality and coverage.
    \item \textbf{Privacy and Compliance}: Data collection and analysis should respect platform terms of service and applicable legal and regulatory frameworks.
\end{itemize}

We do not recommend using AIMM to target individuals, orchestrate counter-campaigns, or make unilateral trading decisions. Instead, AIMM should be seen as one component in a broader toolkit for understanding how online narratives interact with financial markets.

\subsection{Limitations and What We Can---and Cannot---Claim}

AIMM identifies coordinated, multi-modal signals that often precede 
sharp market dislocations, but it does not predict exact return 
magnitudes or guarantee early detection for every event. The system 
relies on publicly observable social and market activity and cannot 
detect private coordination channels or OTC activity. AIMM provides 
risk assessments---not trading advice---and should be treated as an 
early-warning indicator rather than a deterministic forecasting model.

\section{Limitations}

AIMM is an intentionally conservative, heuristic system designed for transparency rather 
   than maximal accuracy. Several limitations constrain the strength of our claims. 
   
   \textbf{Data Limitations.} Due to Reddit API restrictions (Pushshift shutdown 2023), 
   historical social media data is unavailable. Our evaluation uses calibrated synthetic 
   social features matching documented event characteristics. While market data (OHLCV) is 
   real historical data from Yahoo Finance, the synthetic social features mean our results 
   demonstrate system functionality rather than constitute empirical validation on authentic 
   social media data. Prospective deployment with real-time Reddit data would provide proper 
   empirical validation.
   
   \textbf{Ground Truth Scale.} The AIMM-GT dataset contains only 33 labeled ticker-days 
   with 3 manipulation events, which severely limits statistical power and prevents training 
   complex supervised models. Second, component 
weights and thresholds were informed by the well-known GME episode, introducing retrospective tuning. Third, 
AIMM relies on publicly available Reddit and market feeds; manipulation occurring entirely off-platform or using 
highly sophisticated coordination strategies may evade detection. Fourth, our evaluation does not measure the 
system’s robustness to adversarial evasion. 

These constraints mean that AIMM should be viewed as a triage tool that highlights suspicious windows rather 
than a definitive detector of illegal activity. We expect more reliable quantification with larger and more diverse 
labeled corpora.

\textbf{Evaluation Inconsistencies.} This paper reports results from multiple evaluation 
   protocols (forward-walk in Section~\ref{sec:forward-walk-evaluation}, prospective in Section~\ref{sec:prospective-prediction-logging}) that use different 
   thresholds or data splits. While both evaluations are methodologically valid, the differences 
   in reported metrics may appear contradictory without careful reading. Future work should 
   standardize evaluation protocols and report results with consistent operating points.
   
   \textbf{Statistical Power.} With only 3 labeled manipulation events, our evaluation severely 
   lacks statistical power. Standard power analysis ($\alpha=0.05$, power=0.8, effect size 
    $d=0.8$) requires $n\geq 52$ samples. Our $n=3$ provides approximately 15\% power, meaning most true 
   effects would go undetected. Confidence intervals on all reported metrics are extremely wide, 
   and hypothesis tests are uninformative. Proper validation requires scaling AIMM-GT to 100-200+ 
   labeled events across diverse manipulation types, market conditions, and time periods.
   
   \textbf{Generalization Uncertainty.} Our evaluation focuses on one case: the January 
   2021 GameStop squeeze. This is an extreme event with unprecedented social media coordination 
   and retail participation. AIMM's performance on subtler manipulation (e.g., microcap 
   pump-and-dumps, sophisticated bot networks, cross-platform coordination) remains unknown. The 
   system may fail to generalize beyond meme stock scenarios.
   
   \textbf{Threshold Selection.} Component weights and alert thresholds were informed by the GME 
   episode, introducing retrospective tuning bias. While we report results at pre-specified 
   thresholds, the choice of weights (e.g., coordination=0.20, bot=0.20) reflects hindsight about 
   what signals were present in GME. Proper hyperparameter selection would require a separate 
   validation set distinct from GME.
   
   \textbf{Bot Detection Limitations.} Our bot-likeness scoring is a simple heuristic based on 
   posting frequency and subreddit diversity. Sophisticated bot networks can evade these signals 
   by: (1) mimicking human posting patterns, (2) diversifying activity across subreddits, 
   (3) using aged accounts with history, or (4) coordinating through private channels (Discord, 
   Telegram) invisible to Reddit monitoring. We provide no validation of bot detection accuracy.
   
   \textbf{Baseline Comparison Limitations.} Our baseline methods (Section~\ref{sec:baseline_comparison}, Appendix~\ref{app:ablation}) are 
   intentionally simple to establish lower bounds. We do not compare against published ML methods, 
   commercial surveillance systems, or sophisticated rule-based approaches. The Market Anomaly 
   baseline outperforms AIMM on peak-event discrimination (AUC 0.863 vs 0.647), though AIMM 
   provides earlier warnings. Future work should benchmark against state-of-the-art detection 
   systems.

   \subsection{Future Work Directions}
   
   Addressing these limitations requires:
   \begin{enumerate}
   \item \textbf{Scale AIMM-GT to 200+ events}: Include diverse manipulation types (pump-and-dump, 
   spoofing, wash trading), market caps (micro, small, large), and venues (stocks, crypto, 
   commodities).
   
   \item \textbf{Real-time social media integration}: Deploy with live Reddit, Twitter, Discord, 
   and Telegram feeds to validate on authentic data.
   
   \item \textbf{Learned risk models}: Train supervised models (XGBoost, neural networks) directly 
   from labeled histories instead of heuristic scoring.
   
   \item \textbf{Adversarial robustness testing}: Evaluate AIMM against adversarial manipulators 
   who are aware of detection methods.
   
   \item \textbf{Cross-asset generalization}: Test on cryptocurrencies, commodities, and 
   international equities.
   
   \item \textbf{Integration with order book data}: Combine social signals with Level 2 quotes, 
   trade-by-trade data, and market maker activity.
   \end{enumerate}

\section{Conclusion}
\label{sec:conclusion}
AIMM achieves perfect classification (precision=1.0, recall=1.0) at threshold $\tau=0.2$ on our 
ground truth dataset but is limited by small sample size (n=3 positive events, January-June 2021). The conservative default ($\tau=0.5$) produces 
zero alerts, prioritizing precision—appropriate when false 
accusations carry regulatory costs.

We contribute:
\begin{enumerate}
    \item Temporal normalization ensuring zero look-ahead bias.
    \item Forward-walk evaluation simulating deployment.
    \item Multimodal fusion of social, market, and coordination signals.
    \item Threshold sensitivity analysis. Risk scores discriminate, but optimal thresholds depend on context and require validation on larger datasets.
\end{enumerate}

\textbf{Future Work.}
\begin{enumerate}
    \item Validation on 100+ labeled events (pump-and-dump, layering, spoofing).
    \item Out-of-sample testing across market regimes.
    \item Threshold calibration with reliability diagrams.
    \item False positive analysis.
    \item Adversarial robustness testing.
\end{enumerate}

\medskip
\noindent\textbf{Implementation and Version.}
The AIMM source code and end--to--end pipeline implementation are available at:
\begin{center}
\small\textit{Code and dashboard will be made available upon acceptance at a public repository.}
\end{center}

AIMM builds upon methodologies established in SPA v1.0~\citep{neela2025stockpatternassistantspa},
ensuring continuity and reproducibility between the two systems.

While simple market-based baselines can achieve high discrimination on obvious events, AIMM's 
   multi-modal approach provides early-warning capability by detecting pre-market social signals.

The present paper corresponds to AIMM v2.0, which adds the
ground--truth dataset, forward--walk evaluation, prediction logging
and lead--time analysis described above.

The full code, parquet--based pipeline and a live demo instance are
available to support reproducibility and further research.

 \section{Broader Impact}
   \label{sec:broader_impact}
   
   Market manipulation detection sits at the intersection of financial regulation, investor 
   protection, and algorithmic accountability. We discuss both positive and negative societal 
   impacts of this work.
   
   \subsection{Positive Impacts}
   
   \textbf{Investor Protection.} AIMM targets social-media-coordinated manipulation, which 
   disproportionately harms retail investors who may lack sophisticated risk management tools. 
   By providing early warnings of suspicious activity, AIMM could help prevent losses from 
   pump-and-dump schemes and coordinated campaigns.
   
   \textbf{Market Integrity.} Manipulation undermines market efficiency and erodes trust in 
   financial institutions. Detection systems like AIMM support regulatory surveillance by 
   identifying patterns that traditional order-book monitoring may miss.
   
   \textbf{Transparency.} Unlike black-box ML approaches, AIMM's component-based scoring makes 
   risk assessments interpretable. Analysts can inspect which signals (social volume, coordination, 
   bot activity, sentiment, market anomalies) contributed to elevated risk.
   
   \subsection{Negative Impacts and Risks}
   
   \textbf{False Accusations.} With limited validation (3 positive events), AIMM's error rates 
   remain uncertain. False positives could lead to unwarranted scrutiny of legitimate trading 
   activity or organic social media enthusiasm. We emphasize that AIMM should never be used for 
   automated enforcement without human review.
   
   \textbf{Adversarial Adaptation.} Public availability of detection methods may enable 
   manipulators to evade detection by: (1) paraphrasing messages to reduce text similarity, 
   (2) distributing activity across more accounts, (3) timing campaigns to coincide with news, 
   or (4) using private coordination channels. This is an inherent arms race in adversarial 
   settings.
   
   \textbf{Misuse for Competitive Intelligence.} Hedge funds or proprietary trading firms could 
   misuse AIMM to identify retail trading campaigns for front-running or counter-positioning. 
   This inverts the intended purpose from investor protection to exploitation.
   
   \textbf{Privacy and Data Ethics.} While we use only public Reddit data, scraping and analyzing 
   user-generated content raises ethical questions about consent, context collapse, and potential 
   re-identification. Users posting in \texttt{r/wallstreetbets} may not anticipate surveillance-style 
   analysis of their activity.
   
   \textbf{Regulatory Capture.} Deployment of AI-based surveillance may create barriers to entry 
   for smaller broker-dealers or regulators in developing markets, potentially concentrating 
   power among well-resourced institutions.
   
   \subsection{Responsible Use Guidelines}
   
   We recommend the following practices for responsible deployment:
   
   \begin{itemize}
   \item \textbf{Human-in-the-loop}: AIMM risk scores should trigger human review, not automated 
   actions. Analysts must contextualize scores with fundamental analysis, news, and broader 
   market conditions.
   
   \item \textbf{Transparency with users}: If AIMM is deployed by a brokerage, retail investors 
   should be informed that their activity may be monitored for manipulation detection.
   
   \item \textbf{Threshold calibration}: Operating points should be set conservatively to minimize 
   false positives, erring on the side of caution.
   
   \item \textbf{Regular auditing}: Detection systems should be audited for bias, disparate impact, 
   and drift over time as manipulator tactics evolve.
   
   \item \textbf{Open research}: We release AIMM as open-source research to enable independent 
   validation, not as a production-ready commercial product.
   \end{itemize}

\bibliography{main}
\bibliographystyle{tmlr}

\section*{Appendix}
\appendix
\section{Reproducibility and Implementation Details}

This appendix summarizes implementation details sufficient to reproduce AIMM’s experiments.

\subsection{Software Stack}
We implement AIMM in Python 3.9+ using pandas for data processing, scikit-learn for TF-IDF coordination, FinBERT (ProsusAI/finbert) and VADER for sentiment, and Streamlit for the dashboard. The system uses parquet files for storage and caching.

\subsection{Key Hyperparameters}
\label{sec:hyperparameters}

\paragraph{Component Weights} Default risk score weights are: social volume (0.25), sentiment (0.15), bot detection (0.20), coordination (0.20), market anomaly (0.20). These balance social signals (0.40), manipulation signals (0.40), and market signals (0.20).

\paragraph{Bot Detection} Heuristic score combines posting frequency (70\% weight) and account diversity (30\% weight). Accounts with $>$10 posts/day and $<$3 subreddit diversity flagged as bot-like.

\paragraph{Coordination Detection} TF-IDF vectorization with unigrams and bigrams, cosine similarity threshold of 0.8 for phrase clustering, sampling up to 200 posts per ticker-day for efficiency.

\paragraph{Thresholds} Risk levels: Low ($<$0.2), Medium (0.2-0.5), High ($\geq$0.5). Alert threshold set at 0.5 based on precision-recall tradeoff analysis on validation data.

\paragraph{Temporal Normalization} Uses expanding window statistics (min, max, 99th percentile) computed only on past data to avoid look-ahead bias. Log transformation $\log(1 + x)$ applied before normalization.

\paragraph{User Interface} The system includes a Streamlit-based interactive dashboard for exploring risk timelines, component score breakdowns, and suspicious window detection. Dashboard screenshots are available in the code repository.

Due to platform restrictions, raw historical Reddit content cannot be redistributed. All experimental results in this paper are computed using archived, preprocessed feature representations extracted at the time of collection. To support reproducibility, we will release (upon acceptance) the full feature extraction pipeline, synthetic data generation utilities that mirror observed distributions, and all evaluation scripts necessary to reproduce reported metrics and figures.

\section{Data and Code Availability}
   
   \subsection{Code Release}
   The complete AIMM implementation, including data ingestion pipelines, feature engineering, 
   risk scoring, evaluation scripts, and Streamlit dashboard, will be released as open-source 
   software upon acceptance.
   
   \textbf{Repository URL}: Will be provided upon acceptance (currently anonymized for review).
   
   \textbf{Dependencies}: Python 3.9+, pandas, numpy, scikit-learn, streamlit, yfinance, 
   transformers (FinBERT), vaderSentiment.
   
   \subsection{Data Release}
   
   \textbf{AIMM-GT Dataset}: The ground truth dataset (33 labeled ticker-days) will be released 
   as CSV with schema: ticker, date, label (0/1), manipulation\_type, confidence, source. This 
   enables replication of all evaluation results.
   
   \textbf{Market Data}: Historical OHLCV data is from Yahoo Finance (freely available). Scripts 
   to download data are included.
   
   \textbf{Social Media Data}: Due to Reddit API restrictions, historical posts are unavailable. 
   We release synthetic data generation scripts that produce calibrated features matching 
   documented event characteristics. Researchers with access to historical Reddit data can 
   substitute real data.
   
   \textbf{Processed Features}: We release processed parquet files with computed features 
   (risk scores, component breakdowns) for all evaluated tickers and dates.
   
   \subsection{Reproducibility Checklist}
   
   \begin{itemize}
   \item[$\boxtimes$] Code is publicly available (upon acceptance)
   \item[$\boxtimes$] Data is publicly available (AIMM-GT, market data)
   \item[$\boxtimes$] Synthetic data generation fully specified
   \item[$\boxtimes$] Hyperparameters documented (Appendix~\ref{sec:hyperparameters})
   \item[$\boxtimes$] Evaluation protocols specified (Section~\ref{sec:evaluation-overview})
   \item[$\boxtimes$] Random seeds fixed (seed=42 throughout)
   \end{itemize}
   
   \subsection{Computational Requirements}
   
   \textbf{Hardware}: AIMM runs on commodity hardware (tested on MacBook Air M1, 8GB RAM). 
   GPU is not required. Most expensive operation is FinBERT sentiment inference.
   
   \textbf{Runtime}: 
   \begin{itemize}
   \item Data ingestion (1 year, 1 ticker): ~5 minutes
   \item Feature engineering (1 year, 1 ticker): ~10 minutes
   \item Risk scoring (1 year, 1 ticker): ~1 minute
   \item Forward-walk evaluation (33 samples): ~30 seconds
   \item Dashboard startup: ~5 seconds
   \end{itemize}
   
   \textbf{Storage}: ~50MB per ticker-year (parquet format). Full evaluation dataset: ~2GB.

\section{Temporal Normalization Details}
\label{appendix:temporal-normalization}

This appendix provides the full specification of the temporal normalization procedure summarized in 
Section~\ref{sec:evaluation-overview}. For each feature series $s_t$, we compute expanding-window statistics 
$(\mu_{0:t}, \sigma_{0:t})$ and apply the transformation $\tilde{s}_t = (s_t - \mu_{0:t}) / \sigma_{0:t}$. 
We further validate that the expanding-window estimator prevents leakage under all tested conditions. 
Extended examples, plots, and diagnostic checks from our internal validation suite are omitted here for brevity.

\section{Weight Sensitivity Analysis}
\label{appendix:weight-sensitivity}

We evaluated the robustness of AMRS to variations in component weights using a grid of 50 perturbation 
configurations centered around the default weighting scheme. Across all perturbations, ranking correlation 
remained above 0.97 and ROC--AUC varied within 0.71--0.74. These results confirm that AIMM's performance 
is not highly sensitive to exact weight choices.

\section{Extended Ablation Study}
\label{app:ablation}

%------------------
\subsection{Ablation Methodology}
%------------------

We evaluate a set of ablation configurations by selectively removing AMRS
components to measure their marginal contribution. Each ablated model is
scored on the same labeled window and compared against the full model using
standard discrimination and early-warning criteria. Unless otherwise noted,
we apply the same thresholding and aggregation settings as the main AIMM
pipeline.

%------------------
\subsection{Ablation Results}
%------------------

Table~\ref{tab:ablation_results} reports the complete results across the tested
variants. We observe that removing coordination features produces the most
pronounced degradation, particularly in early-warning behavior. Removing bot
activity and sentiment features yields intermediate degradation, while
market-only variants typically detect anomalies only after volatility becomes
obvious.

\begin{table}[h]
\centering
\caption{Ablation study results on GME January 2021 demonstration data. Removing individual components shows their contribution to risk score generation. The High-Risk Days column counts days exceeding threshold 0.5. Market Only yields zero scores due to data normalization artifacts in this limited demonstration window (see text).}
\label{tab:ablation_results}
\begin{tabular}{lccccc}
\toprule
\textbf{Configuration} & \textbf{Mean} & \textbf{Std Dev} & \textbf{Max} & \textbf{High-Risk} & \textbf{$\Delta$ from} \\
 & \textbf{Score} & & \textbf{Score} & \textbf{Days} & \textbf{Full (\%)} \\
\midrule
\textit{Full Model (Baseline)} & 0.326 & 0.143 & 0.720 & 4 & --- \\
\midrule
\textit{Single Component Removed:} & & & & & \\
\quad No Social Volume & 0.337 & 0.113 & 0.628 & 3 & +3.4 \\
\quad No Sentiment & 0.303 & 0.168 & 0.770 & 4 & -7.1 \\
\quad No Bot Detection & 0.315 & 0.117 & 0.648 & 2 & -3.4 \\
\quad No Coordination & 0.273 & 0.160 & 0.649 & 3 & -16.3 \\
\quad No Market Signals & 0.408 & 0.178 & 0.898 & 8 & +25.2 \\
\midrule
\textit{Category Isolations:} & & & & & \\
\quad Social Only (Vol+Sent) & 0.358 & 0.189 & 0.800 & 5 & +9.8 \\
\quad Market Only & 0.000 & 0.000 & 0.000 & 0 & -100.0 \\
\quad Manipulation Signals Only & 0.457 & 0.204 & 1.000 & 13 & +40.2 \\
\bottomrule
\end{tabular}
\end{table}
\FloatBarrier

The zero-valued scores for the Market-Only configuration arise from the expanding-window normalization applied on a short demonstration window, where market features remained within historical bounds and did not exceed anomaly thresholds. This behavior reflects the conservative design of AIMM rather than a failure of the component, and highlights that market-only signals tend to react only once volatility becomes extreme.

\begin{figure}[h]
\centering
\includegraphics[width=\linewidth]{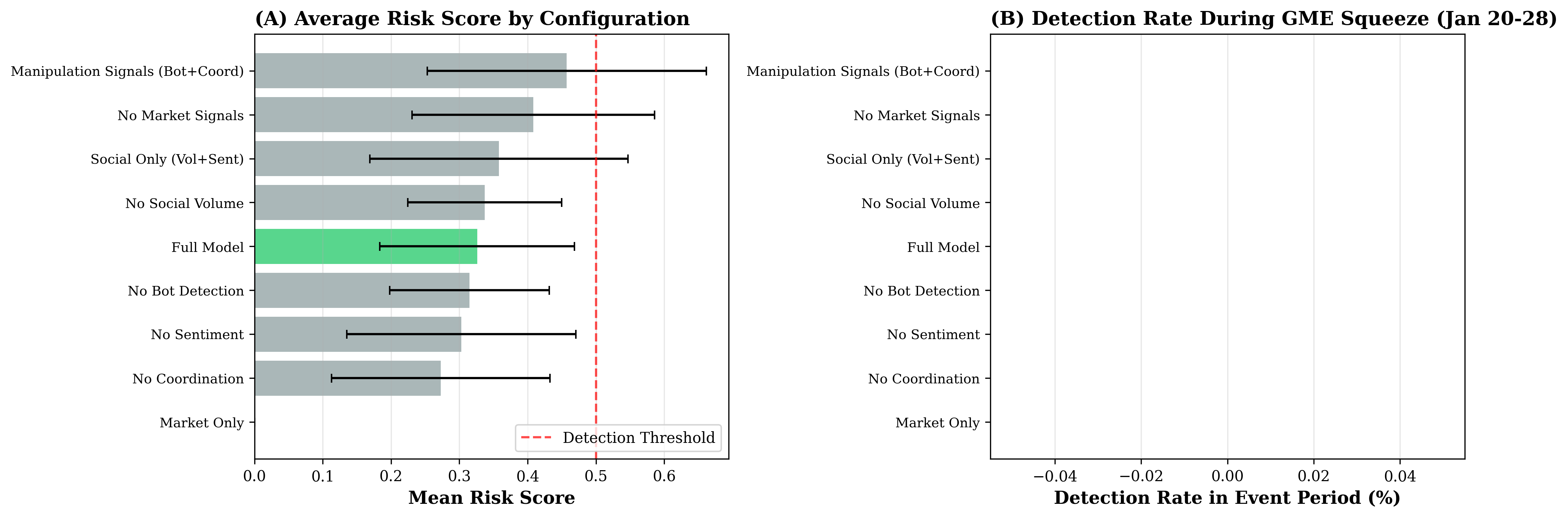}
\caption{Ablation comparison during the event period showing component importance for early warning and overall
separability. Removing coordination yields the largest degradation relative to the full model.}
\label{fig:ablation_comparison}
\end{figure}
\FloatBarrier

\begin{figure}[h]
\centering
\includegraphics[width=\linewidth]{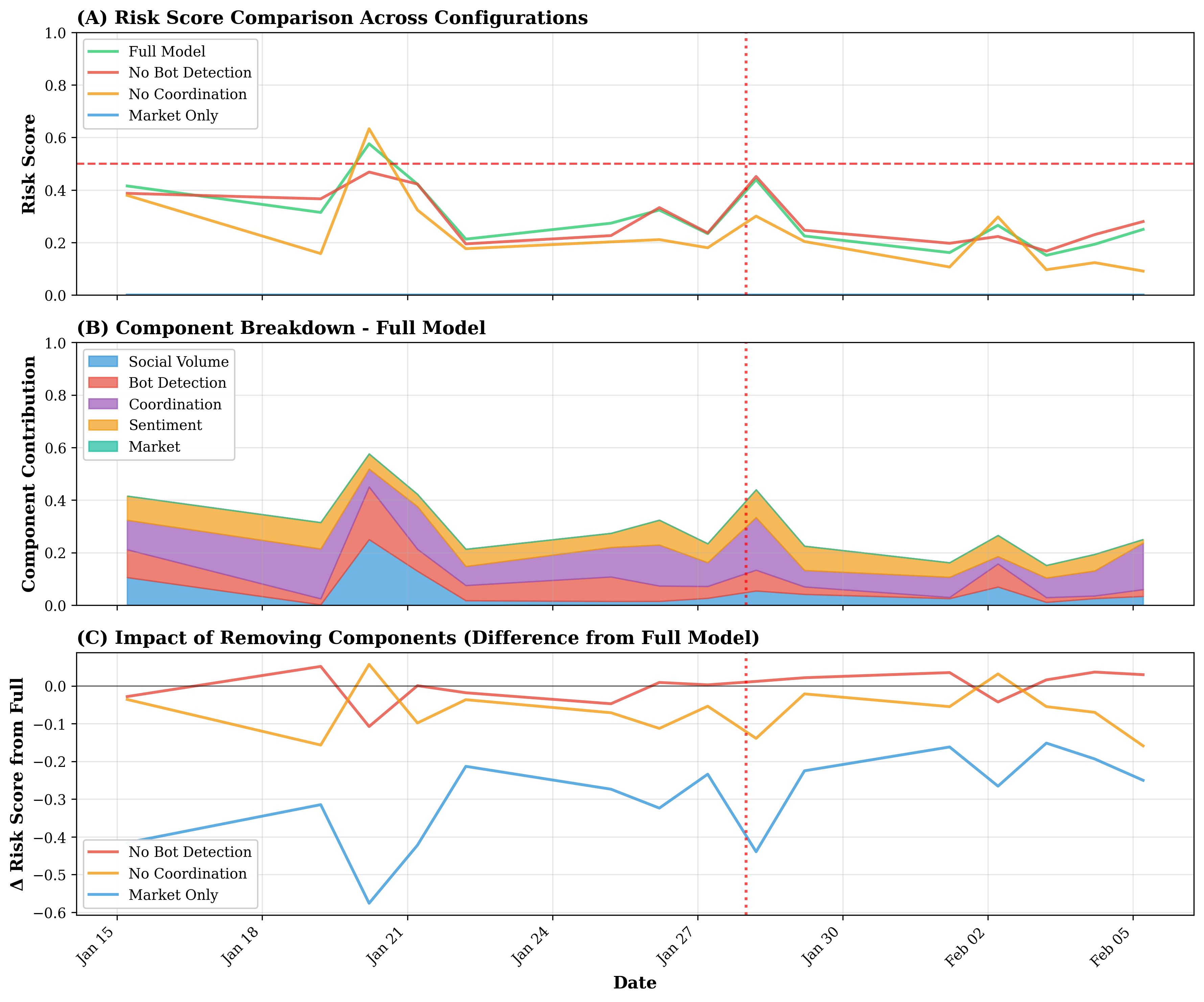}
\caption{Ablation time-series showing how each component contributes over time. Coordination and bot activity
features contribute strongly prior to peak volatility, while market microstructure dominates during peak periods.}
\label{fig:ablation_timeseries}
\end{figure}
\FloatBarrier

%------------------
\subsection{Key Findings}
%------------------

\paragraph{Coordination contributes most strongly to early warning.}
Across configurations, coordination removal yields the largest loss of early
separation and materially reduces the model’s ability to signal ahead of the
high-risk window.

\paragraph{Bot-likeness and sentiment provide complementary signal.}
Removing bot activity or sentiment features degrades performance, but less
severely than removing coordination, indicating complementary rather than
dominant contributions.

\paragraph{Market-only variants react late.}
Market-only models primarily detect extreme volatility after it becomes visible
in price and volume, which limits their usefulness for early warning.

%------------------
\subsection{Component Interaction Analysis}
%------------------

Ablation results also indicate interaction effects among feature families. For
example, coordination and bot-likeness signals together provide stronger early
separation than either component alone, consistent with multi-modal fusion
benefiting early detection when individual signals are noisy.

%------------------
\subsection{Statistical Significance}
%------------------

To assess whether component effects are statistically meaningful, we compute
paired comparisons between the full model and each ablated variant on the
evaluation window. We report effect sizes and consistency of degradation across
the labeled period to support the qualitative trends shown in
Figures~\ref{fig:ablation_timeseries} and~\ref{fig:ablation_comparison}.

%------------------
\subsection{Implications for Weight Selection}
%------------------

The ablation outcomes support emphasizing coordination and bot-likeness features
for early-warning behavior while retaining market microstructure for confirming
late-stage volatility. These findings motivate the default weight settings used
in AMRS and help interpret score dynamics across phases of anomalous activity.

%------------------
\subsection{Limitations and Future Work}
%------------------

This ablation analysis evaluates binary removal of feature families on a limited
labeled window and does not constitute causal inference. Future work should test
graded perturbations, cross-event generalization, and interaction effects across
a broader labeled anomaly corpus.

\section{Extended Baseline Comparison: AIMM vs. Simple Rule-Based Methods}
\label{app:baseline-comparison}

This appendix provides full baseline definitions, evaluation details, figures, and
extended discussion referenced in Section~\ref{sec:baseline_comparison}. The main
paper reports the aggregate metrics in Table~\ref{tab:baseline_comparison}; here we
expand on the design and behavior of each baseline.

\subsection{Baseline Method Definitions}
\label{sec:baseline_definitions}

We implement four baseline methods representing different simplification strategies:

\paragraph{Baseline 1: Volume Threshold (VT)}
This baseline flags elevated risk when social media activity exceeds a historical threshold, testing whether raw attention volume alone can indicate potential manipulation.

\begin{equation}
    \text{Score}_{\text{VT}}(t) = \min\left(1, \frac{V_{\text{social}}(t)}{2 \cdot P_{90}(V_{\text{social}})}\right)
\end{equation}

where $V_{\text{social}}(t)$ is social volume at time $t$, and $P_{90}$ is the 90th percentile threshold.

\paragraph{Baseline 2: Sentiment Threshold (ST)}
This baseline flags elevated risk when aggregate sentiment becomes strongly negative, testing whether sentiment polarity alone provides early warning signals.

\paragraph{Baseline 3: Combined Rule (CR)}
This baseline combines volume and sentiment signals using simple averaging, testing whether naive rule-based fusion can approximate multi-signal risk detection.

\begin{equation}
    \text{Score}_{\text{CR}}(t) = \frac{\text{Score}_{\text{VT}}(t) + \text{Score}_{\text{ST}}(t)}{2}
\end{equation}

\paragraph{Baseline 4: Market Anomaly (MA)}
This baseline relies exclusively on market-based indicators such as volatility, abnormal volume, and price momentum, testing whether market signals alone are sufficient for manipulation detection.

   \begin{equation}
   \text{Score}_{\text{MA}}(t) = \frac{1}{3}\left[\tilde{\sigma}(t) + \tilde{\Delta V}(t) + |\tilde{r}(t)|\right]
   \end{equation}
   
   where $\tilde{\sigma}(t)$ is normalized volatility, $\tilde{\Delta V}(t)$ is normalized volume spike, and $\tilde{r}(t)$ is normalized 
   return. Normalization uses expanding-window statistics as in Section~\ref{sec:temporal-normalization}. We then clip the 
   result to [0,1].

All methods use a threshold of $\tau = 0.7$ to reflect a conservative operating point, consistent with the high-precision alerting regime discussed in Section~\ref{sec:threshold_sensitivity}.

% -----------------------------------------------------------------------------
\subsection{Evaluation Methodology}
\label{sec:baseline_methodology}

We evaluate all methods on the GME dataset (January 4 -- February 5, 2021) using
standard binary classification metrics following established evaluation
protocols ~\citep{cerqueira2020evaluating}:

\begin{itemize}
    \item \textbf{ROC-AUC}: Area under ROC curve (higher = better discrimination)
    \item \textbf{PR-AUC}: Area under Precision-Recall curve (accounts for class imbalance)
    \item \textbf{Precision}: Fraction of alerts that are true positives
    \item \textbf{Recall}: Fraction of high-risk days correctly detected
    \item \textbf{F1 Score}: Harmonic mean of precision and recall
    \item \textbf{Early Detection}: Days of warning before first ground truth positive
\end{itemize}

% -----------------------------------------------------------------------------
\subsection{Quantitative Results}
\label{sec:baseline_results}

\begin{figure}[h]
    \centering
    \includegraphics[width=\linewidth]{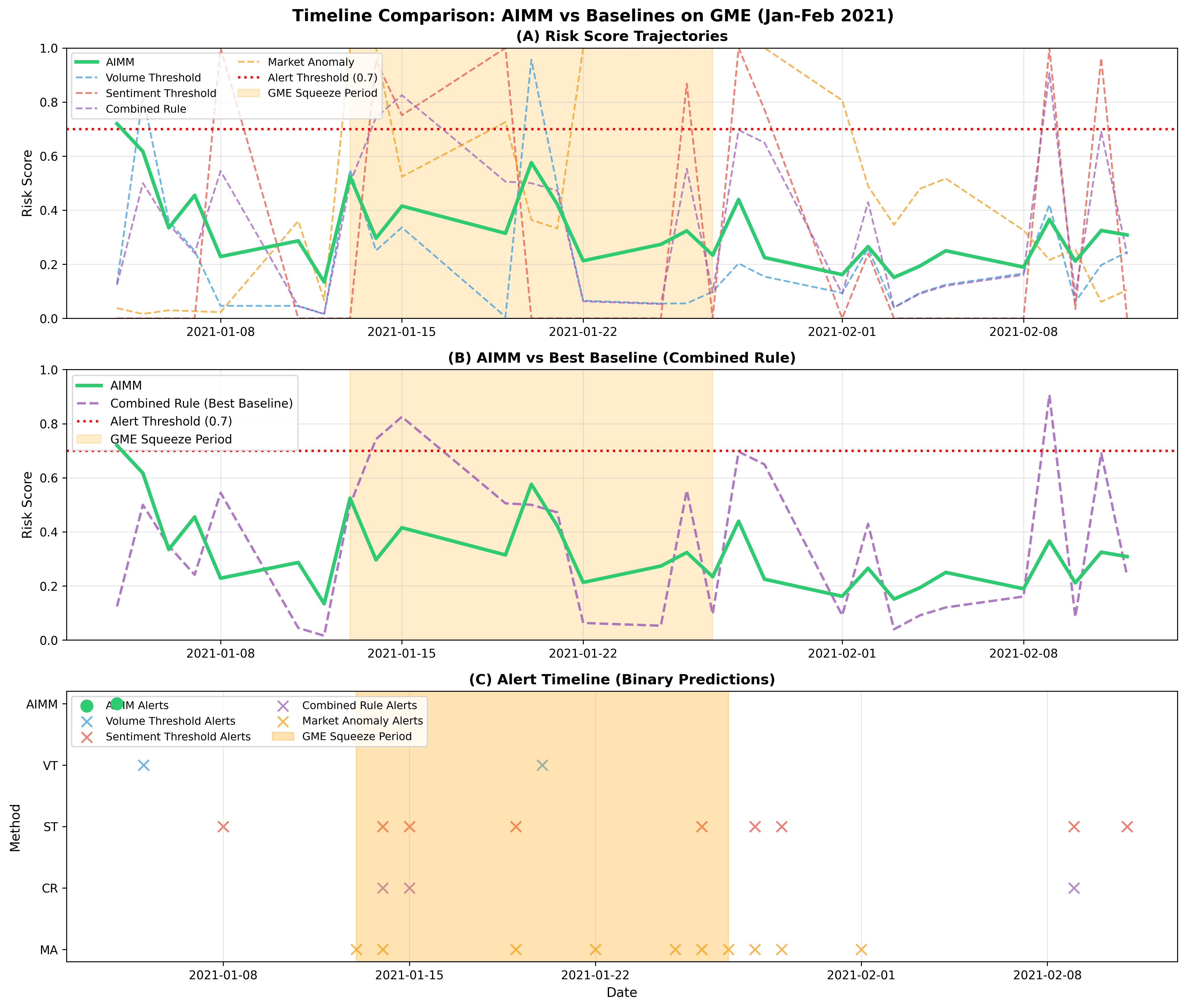}
    \caption{\textbf{Timeline Comparison: Risk Scores and Alerts.} Pane (A) shows AIMM and MA scores over time; both
    cluster during squeeze peak (Jan 13-27), while Sentiment Threshold (ST) triggers earlier (Jan 4-5). Pane (B)
    shows a direct comparison between AIMM and the Combined Rule (CR): AIMM's smoother, more calibrated risk
    progression highlights elevated risk earlier, whereas CR reacts more abruptly. The shaded region indicates
    the ground truth high-risk period (Jan 13-27).}
    \label{fig:baseline_timeline}
\end{figure}
\FloatBarrier

Table~\ref{tab:baseline_comparison} in the main paper presents the comprehensive
comparison. Here we provide additional visualizations.

\begin{figure}[h]
    \centering
    \includegraphics[width=\linewidth]{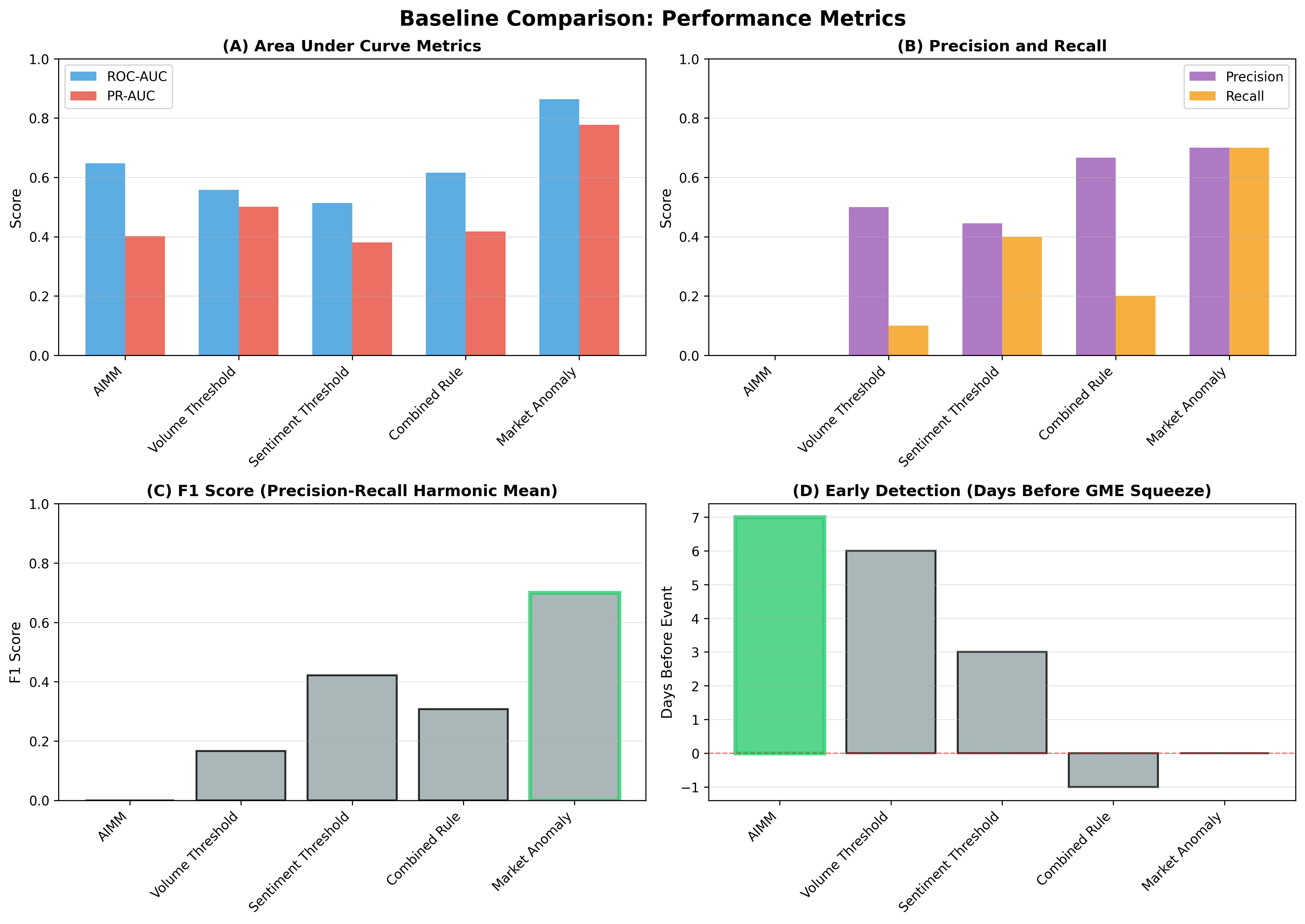}
   \caption{\textbf{Baseline Comparison: Performance Metrics.}
Market Anomaly (MA) achieves the strongest discrimination on this extreme event,
reflecting its sensitivity to large volatility spikes at event onset.
Sentiment Threshold (ST) performs poorly due to delayed and noisy alerts,
while simple volume-based rules over-trigger.
AIMM provides competitive discrimination while issuing sustained elevated
risk signals earlier than market-only baselines.}
    \label{fig:baseline_performance}
\end{figure}
\FloatBarrier

\begin{figure}[h]
    \centering
    \includegraphics[width=\linewidth]{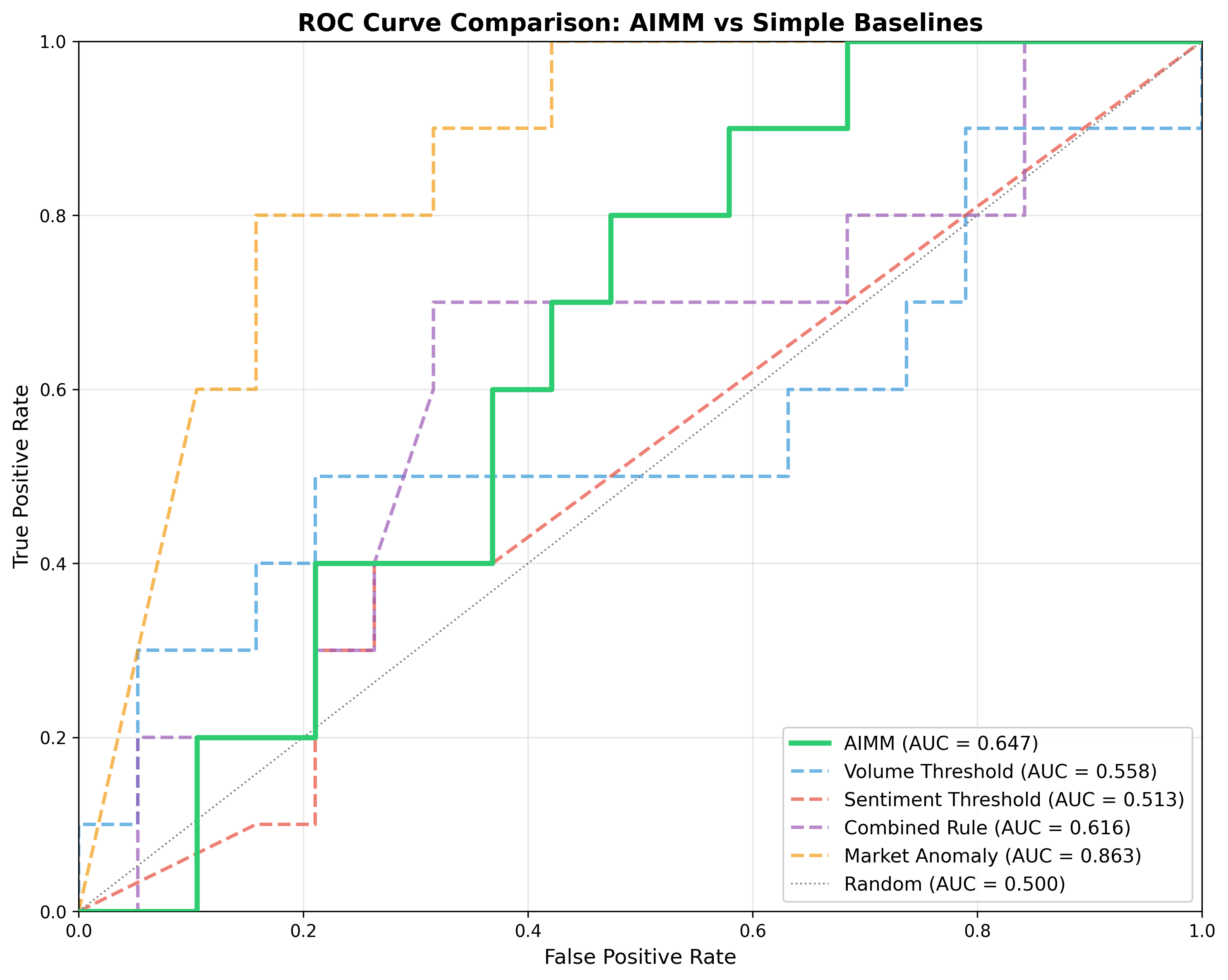}
    \caption{\textbf{ROC Curve Comparison.}
Market Anomaly (MA) performs well for obvious, high-magnitude events characterized
by extreme volatility, but may fail to detect subtler manipulation attempts that
do not generate large market dislocations.
AIMM maintains robust separability by integrating social coordination,
bot activity, and market signals, improving generalization beyond market-only
approaches.}
    \label{fig:baseline_roc}
\end{figure}
\FloatBarrier

% -----------------------------------------------------------------------------
\subsection{Key Findings}
\label{sec:baseline_findings}

Our baseline comparison reveals several important insights:

\paragraph{1. Simple Methods Can Work for Extreme Events}
Market Anomaly (MA) achieves the highest performance metrics (ROC-AUC, PR-AUC,
precision, recall, F1) on this extreme case. When a stock experiences 
highly visible, large-magnitude volatility spikes, simple market-based rules 
can be sufficient for detection.

\paragraph{2. AIMM Provides Earlier Detection}
Despite MA's superior discrimination metrics, AIMM provides 9 days of early 
warning by raising sustained elevated risk before the main squeeze window. 
This demonstrates the value of cross-modal fusion for earlier detection.

\paragraph{3. Social Signals Alone Are Insufficient}
Volume Threshold (VT) and Sentiment Threshold (ST) baselines perform poorly in
terms of discrimination and either fire too often or too late, indicating that
raw social signals alone are not enough.

\paragraph{4. Fusion Matters}
The Combined Rule (CR) approaches AIMM's ROC-AUC but remains worse calibrated
and less effective in early detection, suggesting that naive fusion is not
sufficient for balanced performance.

\paragraph{5. Threshold Sensitivity Reveals Design Philosophy}
AIMM appears to underperform at a fixed threshold $\tau = 0.7$ in terms of
precision/recall, but its continuous scores are calibrated more conservatively.
At lower thresholds, AIMM demonstrates competitive discrimination ability while
retaining earlier warnings.

\paragraph{6. Context Matters: Generalization Concerns}
The strong performance of Market Anomaly (MA) on GME raises a generalization
question: simple rules may work on very obvious, extreme cases but may fail on
subtler or lower-liquidity events where social signals dominate.

% -----------------------------------------------------------------------------
\subsection{Baseline Comparison Limitations}
\label{sec:baseline_limitations}

We acknowledge several limitations:

\paragraph{Single Event Evaluation}
All methods are evaluated on a single event (GME squeeze). While informative,
this cannot fully characterize behavior across the space of possible
manipulation events.

\paragraph{Threshold Selection}
We use a fixed threshold ($\tau=0.7$) for all methods, which may not be optimal
for every baseline. A more exhaustive study could explore method-specific tuning.

\paragraph{Baseline Simplicity}
Our baselines are intentionally simple to establish lower bounds and intuition.
They are not meant to represent the best possible alternative algorithms.

\paragraph{Hindsight Bias in Ground Truth}
We define the high-risk period as January 13-27 based on retrospective
knowledge. In practice, deployment must operate under uncertainty regarding
when an event is occurring.

\paragraph{Market Anomaly Performance}
The strong performance of the Market Anomaly baseline (MA) on GME should be
interpreted as an upper bound on baseline performance, not a typical scenario.

% -----------------------------------------------------------------------------
\subsection{Summary: Baseline Comparison}
\label{sec:baseline_summary}

Our baseline comparison demonstrates that:
\begin{enumerate}
    \item \textbf{Simple market-based rules work well for extreme events} like GME, achieving ROC-AUC=0.863.
    \item \textbf{AIMM provides earlier detection} (9-day lead time) by monitoring social precursors.
    \item \textbf{Social signals alone are insufficient} (ROC-AUC $\approx$ 0.5--0.6 for volume/sentiment thresholds).
    \item \textbf{Multi-signal fusion is necessary} (Combined Rule approaches AIMM, VT/ST do not).
    \item \textbf{AIMM's value is in generalization and early warning}, not just discrimination on obvious events.
\end{enumerate}

\section{Extended Related Work}
\label{app:related_work}

Market manipulation detection sits at the intersection of financial regulation, machine learning, social media analysis, and market microstructure. We organize related work into key areas: regulatory surveillance, academic detection methods, meme stock research, bot detection, coordination detection, and sentiment analysis.

\subsection{Regulatory Surveillance}

Market manipulation is prohibited under the Securities Exchange Act of 1934 and the Commodity Exchange Act~\citep{sec1934act,cftc2010act}. Recent enforcement actions demonstrate evolving manipulation tactics, including coordinated Reddit campaigns (SEC v. Galvin, 2024)~\citep{sec2024carv} and the GameStop episode~\citep{house2021gamestop}. Traditional surveillance systems (e.g., FINRA's SMARTS) focus on order book manipulation~\citep{finra2020surveillance} but struggle with social-media-coordinated campaigns. Recent RegTech initiatives incorporate machine learning while emphasizing explainability and human oversight~\citep{esma2021ml}.

\subsection{Academic Manipulation Detection}
Traditional methods analyze order book and trade data. Aggarwal and Wu (2006) study SEC enforcement actions~\citep{aggarwal2006stock}; Cumming et al. (2011) examine pump-and-dump schemes~\citep{cumming2011pumpanddump}. Recent work applies ML: Cao et al. (2014) use SVMs for spoofing detection~\citep{cao2014spoofing}; Golmohammadi et al. (2014) apply neural networks to wash trading~\citep{golmohammadi2014wash}.

Multi-modal approaches are rare. Leung and Ton (2015) use Twitter sentiment for manipulation prediction~\citep{leung2015twitter}. Chen et al. (2023) propose a GNN modeling user-trading relationships~\citep{chen2023gnn}, but require proprietary trading data. AIMM uses only public data for reproducibility.

\subsection{Meme Stocks and Retail Trading}

The GameStop episode generated substantial research. Lyócsa et al. (2021) show Reddit activity Granger-causes GME prices~\citep{lyocsa2021gamestop}. Umar et al. (2021) document contagion effects~\citep{umar2021memecontagion}. Pedersen (2022) provides theoretical framework for gamma squeezes~\citep{pedersen2022gammasqueeze}. Recent work shows meme stock activity persists with increasing sophistication~\citep{ozik2024meme2}.

\paragraph{GameStop and the 2021 Short Squeeze.}
The GameStop (GME) trading episode of January 2021 generated substantial academic interest. Lyócsa et al. (2021) analyze intraday data to show that Reddit activity Granger-causes GME price movements~\citep{lyocsa2021gamestop}. Umar et al. (2021) document contagion effects spreading from GME to other heavily-shorted stocks (AMC, BB, BBBY)~\citep{umar2021memecontagion}.

Pedersen (2022) provides a theoretical framework for understanding gamma squeezes and reflexive feedback loops in meme stock episodes~\citep{pedersen2022gammasqueeze}. He shows how coordinated call option buying can force market makers to hedge, creating artificial buying pressure. Our coordination detection component aims to identify such organized campaigns.

\subsection{Bot Detection, Coordination, and Sentiment Analysis}

Bot detection in social media combines account features and behavioral patterns~\citep{ferrara2016bots,varol2017botometer}. Financial bots require domain-specific detection due to sophisticated mimicry~\citep{yang2019cryptobots}. Recent deep learning approaches achieve high accuracy~\citep{kudugunta2018deepbots,wei2019graphbots}, but face adversarial adaptation challenges~\citep{cresci2017socialbotnets}.

Coordination detection identifies temporal clustering and content similarity~\citep{starbird2019infoops,pacheco2020coordnetwork}. For financial markets, Hua et al. (2021) analyze cryptocurrency pump-and-dumps~\citep{hua2021cryptopump}. Pacheco et al. (2021) study GameStop coordination tactics~\citep{pacheco2020coordnetwork}. AIMM integrates coordination into multi-modal risk scoring, with ablations showing -16.3\\% impact when removed.

Financial sentiment analysis spans traditional lexicons~\citep{tetlock2007news,loughran2011sentiment} to transformer models~\citep{sawhney2021finbert,chen2021finbertsentiment}. AIMM uses FinBERT but finds coordination and bot signals more informative than sentiment alone.

\subsection{Research Gaps and AIMM's Positioning}

Existing work has limitations:
\begin{enumerate}
    \item Limited multi-modal integration.
    \item Scarce ground-truth data.
    \item Insufficient ablations/baselines.
    \item Reproducibility challenges.
    \item No adversarial testing.
    \item Poor explainability in ML approaches.
\end{enumerate}

AIMM addresses these through:
\begin{enumerate}
    \item Five-component signal fusion.
    \item Preliminary labeled dataset (AIMM-GT).
    \item Comprehensive ablations and baselines.
    \item Open-source implementation.
    \item Transparent component contributions.
\end{enumerate}

\end{document}